\pdfoutput=1
\documentclass[11pt]{article}

\usepackage[]{acl}

\usepackage{times}
\usepackage{latexsym}

\usepackage[T1]{fontenc}

\usepackage[utf8]{inputenc}

\usepackage{subfigure}
\usepackage{todonotes}

\usepackage{microtype}
\usepackage{booktabs}
\usepackage{multirow}

\usepackage{amssymb}
\usepackage{times}
\usepackage{tabu}
\usepackage{latexsym}
\usepackage{linguex}
\usepackage{graphicx}
\usepackage{booktabs}
\usepackage{multirow, makecell}
\usepackage{color, colortbl}

\usepackage{etoolbox}
\usepackage{numprint}
\patchcmd{\quote}{\rightmargin}{\leftmargin 1em \rightmargin}{}{}
\usepackage{multirow}
\usepackage{microtype}
\usepackage{tikz}
\usepackage{amssymb}
\usepackage{mdframed}
\usepackage{color}
\usepackage{xcolor}
\usepackage{colortbl}
\usepackage[normalem]{ulem}
\usepackage{amsthm}
\usepackage{cleveref}

\definecolor{emerald}{rgb}{0.31, 0.78, 0.47}
\definecolor{lightcoral}{rgb}{0.94, 0.5, 0.5}
\definecolor{gold(web)(golden)}{rgb}{1.0, 0.84, 0.0}
\definecolor{lightcornflowerblue}{rgb}{0.6, 0.81, 0.93}
\definecolor{Gray}{gray}{0.9}
\definecolor{aquamarine}{rgb}{0.5, 1.0, 0.83}
\definecolor{atomictangerine}{rgb}{1.0, 0.6, 0.4}
 \definecolor{babypink}{rgb}{0.96, 0.76, 0.76}

\newcolumntype{L}[1]{>{\raggedright\let\newline\\\arraybackslash\hspace{0pt}}p{#1}}

\title{On the Origin of Hallucinations in Conversational Models:\\ Is it the Datasets or the Models?}

\author{
  Nouha Dziri$^{\dagger\:\lozenge}$  \quad Sivan Milton${^\ddagger}$   \quad {\bf Mo Yu}$^{\P}$  \quad Osmar Zaiane$^{\dagger}$ 
 \quad {\bf Siva Reddy}$^{\lozenge\:\ddagger}$ \\
  $^\dagger$University of Alberta \quad $^\lozenge$Mila -- Quebec AI Institute \\ 
  \quad $^\ddagger$McGill University \quad $^\P$IBM Research \\
  \texttt{dziri@cs.ualberta.ca}
  }

\begin{document}
\maketitle
\begin{abstract}
Knowledge-grounded conversational models are known to suffer from producing factually invalid statements, a phenomenon commonly called hallucination. 
In this work, we investigate the underlying causes of this phenomenon: is hallucination due to the training data, or to the models? 
We conduct a comprehensive human study on both existing knowledge-grounded conversational benchmarks and several state-of-the-art models.
Our study reveals that the standard benchmarks consist of >$60\%$ hallucinated responses, leading to models that not only hallucinate but even amplify hallucinations. Our findings raise important questions on the quality of existing datasets and models trained using them.
We make our annotations publicly available for future research.\footnote{\href{https://github.com/McGill-NLP/FaithDial}{https://github.com/McGill-NLP/FaithDial}} 

\end{abstract}

\section{Introduction}

Knowledge-grounded conversational models, powered by large pre-trained language models \citep{radford2019language, NEURIPS2020_1457c0d6, 2020t5}, are well-known to generate factually incorrect statements, a phenomenon commonly called \textit{hallucination}  \cite{dziri2021evaluating, rashkin-etal-2021-increasing}. 
A large commonality in the majority of prior work seeks to address hallucination by ameliorating the model \cite{shuster-etal-2021-retrieval-augmentation, mielke2020linguistic, dziri-etal-2021-neural, rashkin-etal-2021-increasing}, but no attempt has been made so far to audit the conversational benchmarks to the best of our knowledge.

\begin{figure}[t]
    \centering
    \includegraphics[width=0.9\linewidth]{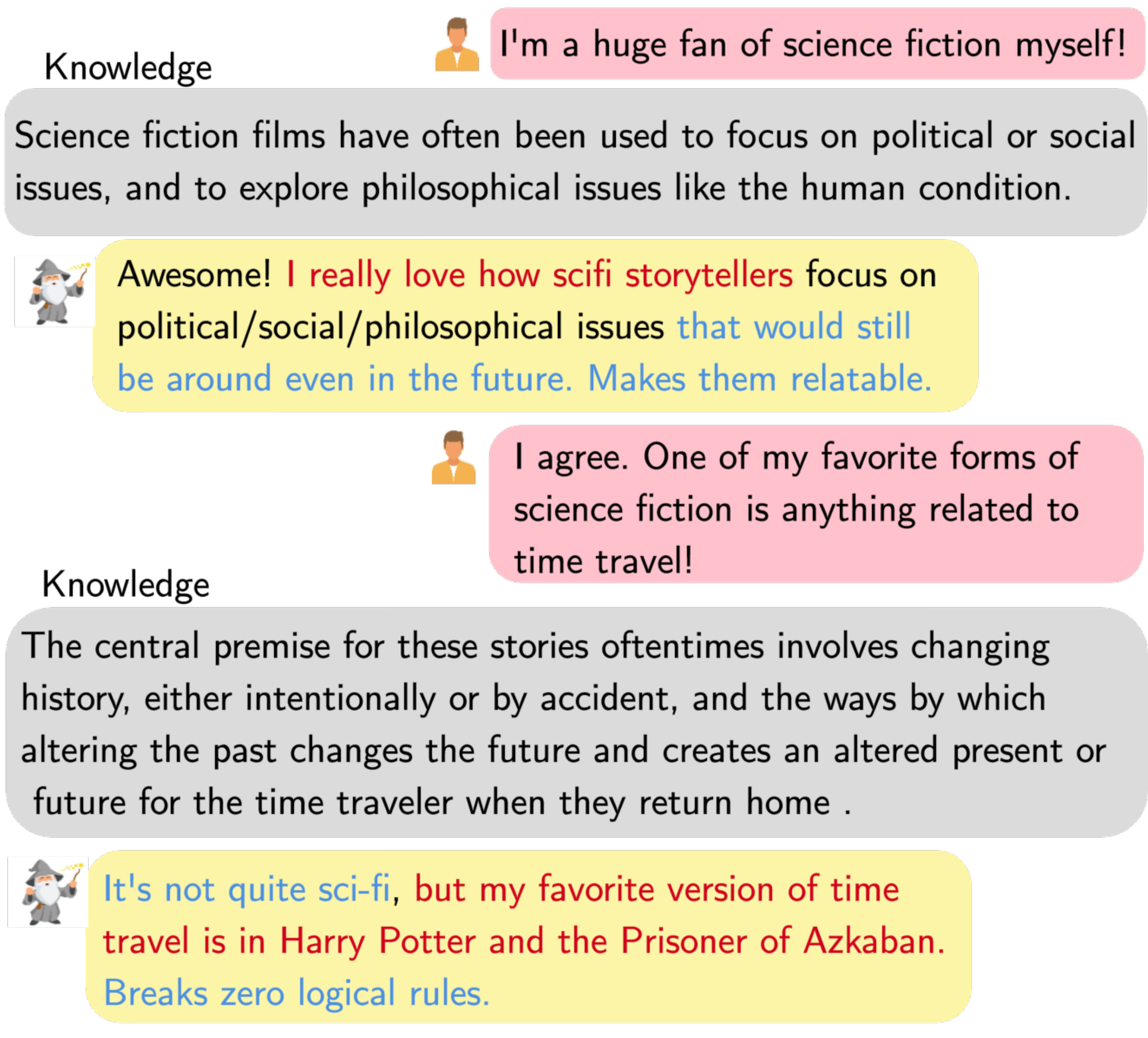}
    \caption{\small{An example of a hallucinated conversation from the Wizard of Wikipedia dataset \cite{dinan2018wizard}. The wizard (yellow) is hallucinating information that cannot be inferred from the knowledge-snippet: hallucinated subjective content (\textcolor{red}{red}) and hallucinated objective content (\textcolor{blue}{blue}).}}
    \label{fig:hallu_dialog}
    \vspace{-15pt}
\end{figure}

On one hand, knowledge-grounded conversational benchmarks may contain hallucinations due to error-prone collection protocols,
or due to a design framework that encourages informativeness over faithfulness.
Existing dialogue systems are typically trained on corpora crowd-sourced through online platforms \cite{dinan2018wizard, Gopalakrishnan2019, moon-etal-2019-opendialkg}.
With loose incentive to come up with faithfully-grounded utterances on the provided knowledge,
crowdworkers may ignore knowledge-snippets altogether, use their personal knowledge or sometimes assume a fictional persona, resulting in conversations that are rife with subjective content and unverified factual knowledge.
Figure~\ref{fig:hallu_dialog} shows a hallucinated conversation from the \textsc{WoW} dataset \cite{dinan2018wizard},

On the other hand, neural conversational models are not necessarily designed to generate faithful outputs, but to mimic the distributional properties of the data.
This kind of optimization will likely push the models to replicate and even amplify the hallucination behaviour at test time \cite{bender2021dangers}. The presence of even few hallucinated responses may skew the data distribution in a way that curbs the model's ability to generate faithful responses \cite{kang-hashimoto-2020-improved}.

In this work, 
drawing insights from the linguistic coding system for discourse phenomena \cite{stiles1992describing} and evaluation frameworks such as  BEGIN \cite{dziri2021evaluating}
and AIS \cite{rashkin2021measuring}, %
we annotate  responses from the three widely-used knowledge-grounded conversational benchmarks: Wizard of Wikipedia \cite{dinan2018wizard}, \textsc{CMU-DoG} \cite{zhou-etal-2018-dataset} and \textsc{TopicalChat} \cite{Gopalakrishnan2019}. 

Our analysis reveals surprisingly that more than 60\% of the responses are hallucinated in the three datasets, with major hallucination modes that manifest principally through the expression of subjective information (e.g., thoughts, beliefs, feelings, intentions, personal experiences) and the expression of unsupported objective factual information.
Further, to understand if neural conversational models make this hallucination more severe, we annotate responses generated by several state-of-the-art models, including ones that are designed to alleviate hallucinations. 
We find that the generated responses consist of an even larger portion of hallucinations, in comparison with  the training data. 
Our findings question the quality of current conversational datasets, their appropriateness to train knowledge-grounded conversational systems, and the robustness of existing models.

\section{Hallucinations in Benchmarks}
\label{sec2}

We conduct a human study on three English crowdsourced  knowledge-grounded conversational benchmarks:
Wizard of Wikipedia (\textsc{WoW}), \textsc{CMU-DoG} and \textsc{TopicalChat}. These datasets consist of dialogues between two speakers, where the goal is to communicate information about particular topics while speakers are presented with a knowledge snippet relevant to the current turn.
More details about these datasets are provided in~\S\ref{appendix:dataset}.

\begin{figure*}
\centering
    \vspace{-2em}
     \subfigure[\label{fig:wizard_begin}\footnotesize Expert annotations (200 responses)]{\includegraphics[width=0.42
     \linewidth]{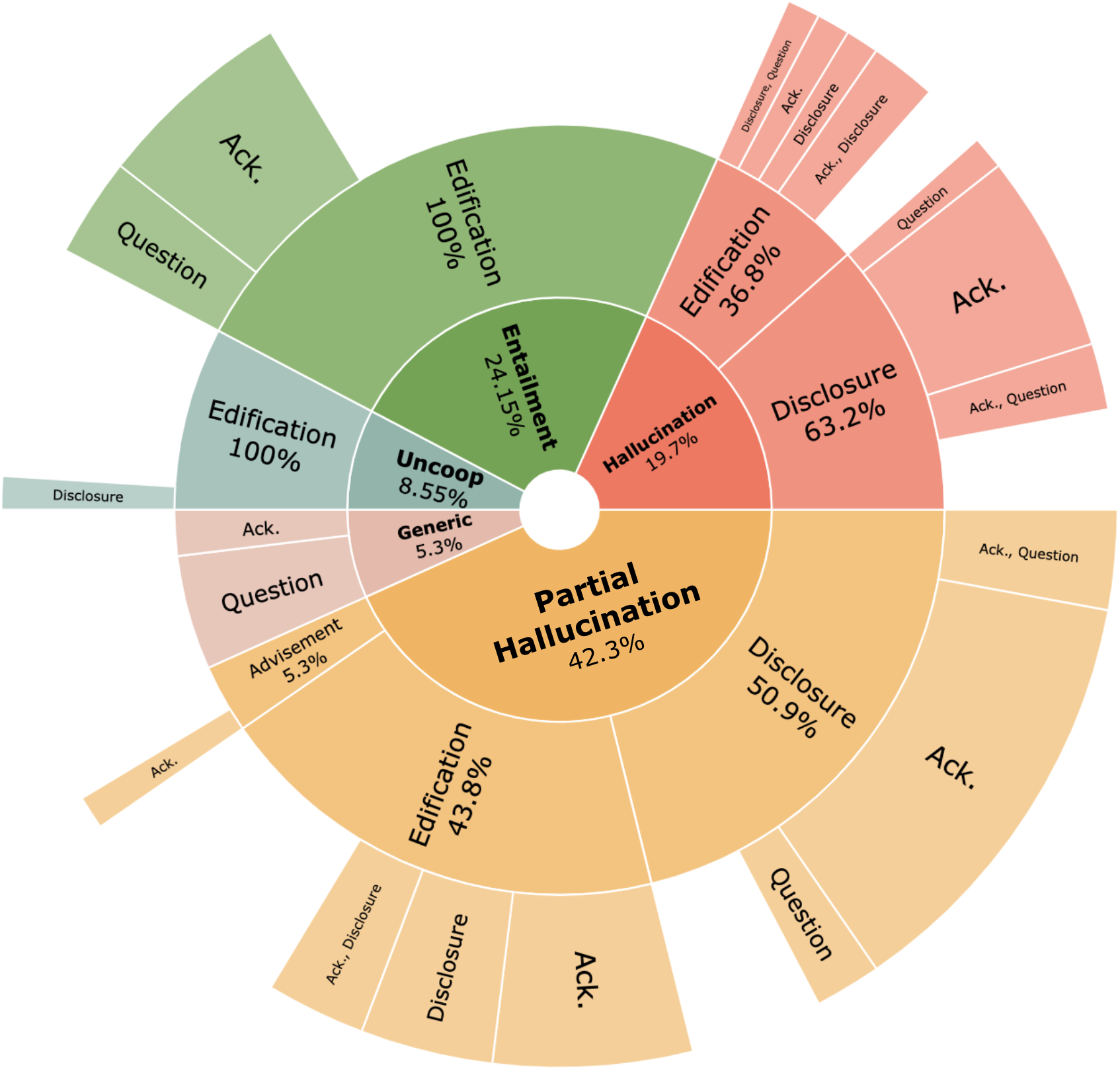}}
    \subfigure[\label{fig:cmu_begin_2}\footnotesize Non-expert annotations (4000 responses)]{\includegraphics[width=0.42\linewidth]{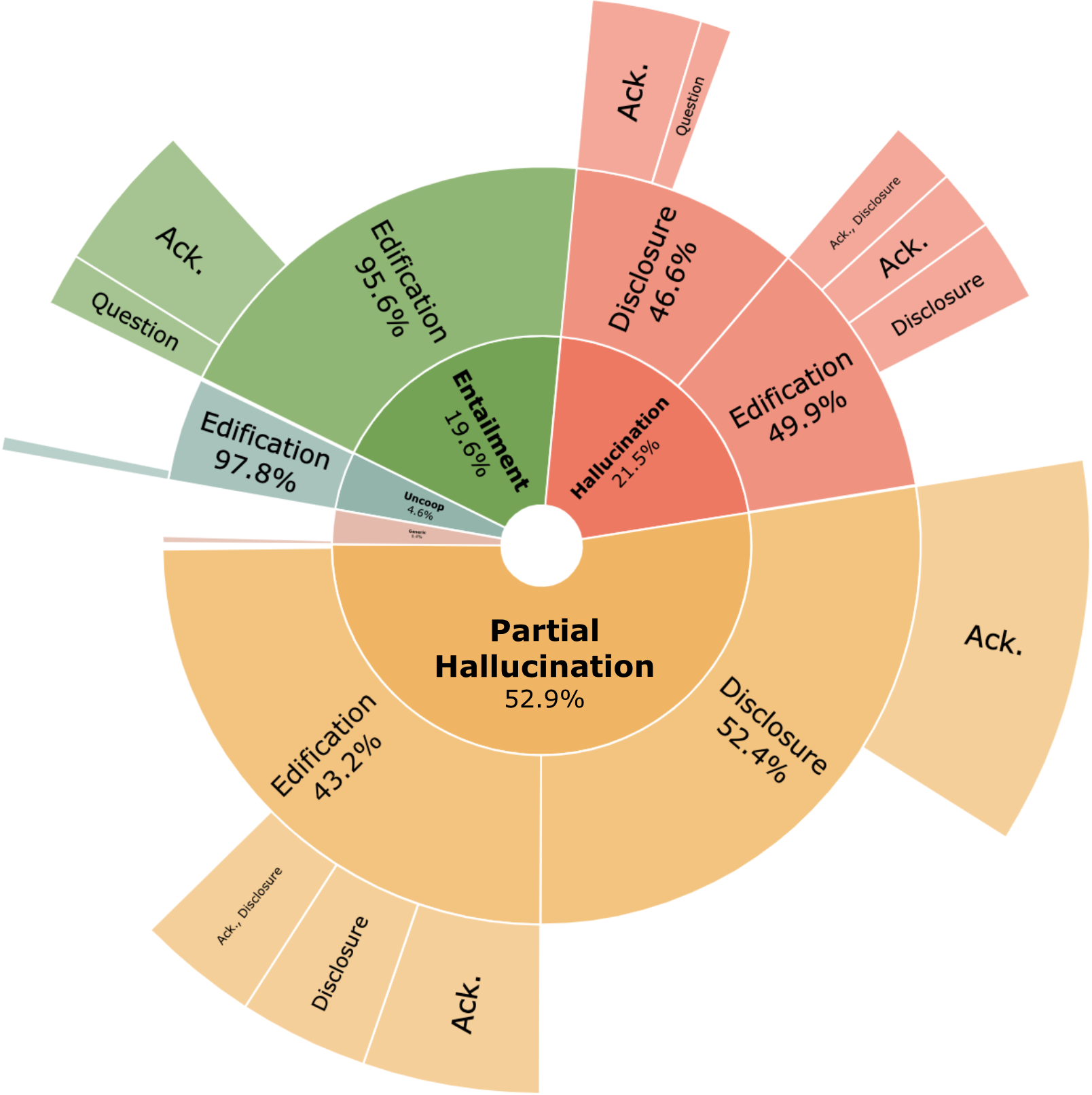}}
    \caption{\small BEGIN and VRM breakdown of responses from WoW. The inner circle shows the breakdown of BEGIN classes and the outer shows the VRM types in each BEGIN type: Hallucination (red), Entailment (green), Partial Hallucination (yellow), Generic (pink), and Uncooperative (blue).}
    \label{fig:breakdown_gold}
\end{figure*} 

\paragraph{Response Classification Taxonomy}

Following the definitions of the BEGIN taxonomy \citep{dziri2021evaluating} 
and the AIS framework \citep{rashkin2021measuring} of evaluating response attribution,
we annotate each response based on whether it can be inferred exclusively from the knowledge-snippet as follows:
\textbf{Entailment}: a response is fully supported by the knowledge, i.e., any information it contains must be attributed to the knowledge.
\textbf{Hallucination}: a response's factual correctness cannot be fully verified from the knowledge-snippet (even if it is true in the real world).  More specifically, personal opinions, experiences, feelings, internal assessments of reality that cannot be attributed to  the information present in the source document, are considered hallucinations. 
\textbf{Partial Hallucination}: part of the response is hallucinated while the rest is entailed by the source knowledge.
\textbf{Generic}: a response that is vague and does not convey any factual information such as ``\textit{Sounds good}" or ``\textit{I'm not sure about that}".
\textbf{Uncooperative}: an entailed response that does not follow the principles of conversational cooperation according to Gricean maxims \cite{grice1989studies}.   The response may be purposefully misleading, or showing a general unwillingness to cooperate with the interlocutor, resulting in an incoherent communication.

To understand the linguistic nature of hallucinations, we further annotate responses based on a linguistic coding system for discourse phenomena, dubbed Verbal Response Modes (VRM; \citealt{stiles1992describing}). 
Concretely, we label a turn with the following speech acts: \textbf{Disclosure}, \textbf{Edification}, \textbf{Advisement}, \textbf{Confirmation}, \textbf{Question} and \textbf{Acknowledgement (Ack.)}. 
Table \ref{fig:vrm_definitions} displays the definition for each \textsc{VRM} type. We opted for the \textsc{VRM} taxonomy as it offers a simple way of codifying responses into categories that are sufficient for our analysis whereas one can also opt for a more demanding annotation scheme \cite{bunt-etal-2020-iso}.
\subsection{Human Evaluation Study}
We follow a two-stage annotation protocol where
 we first ask two linguists to judge the attribution of 200 randomly sampled train responses with respect to the source knowledge. Details about experts can be found in~\S\ref{expert}.  For inter-annotator agreement, we measure Fleiss' Kappa scores on both \textsc{BEGIN} and \textsc{VRM}. \textsc{WoW} achieved 0.89 on BEGIN and 0.78 on \textsc{VRM}, indicating substantial agreement.  Annotations on  \textsc{CMU-DoG} and \textsc{TopicalChat} achieved  nearly similar agreement (See~\S\ref{kappa}). The high agreement scores align with the findings in AIS on \textsc{WoW} \cite{rashkin2021measuring}.
 
\begin{figure*}
    \vspace{-2em}
    \subfigure[\label{fig:cmu_begin_1}\footnotesize CMU-DoG responses]{\includegraphics[width=0.48\linewidth]{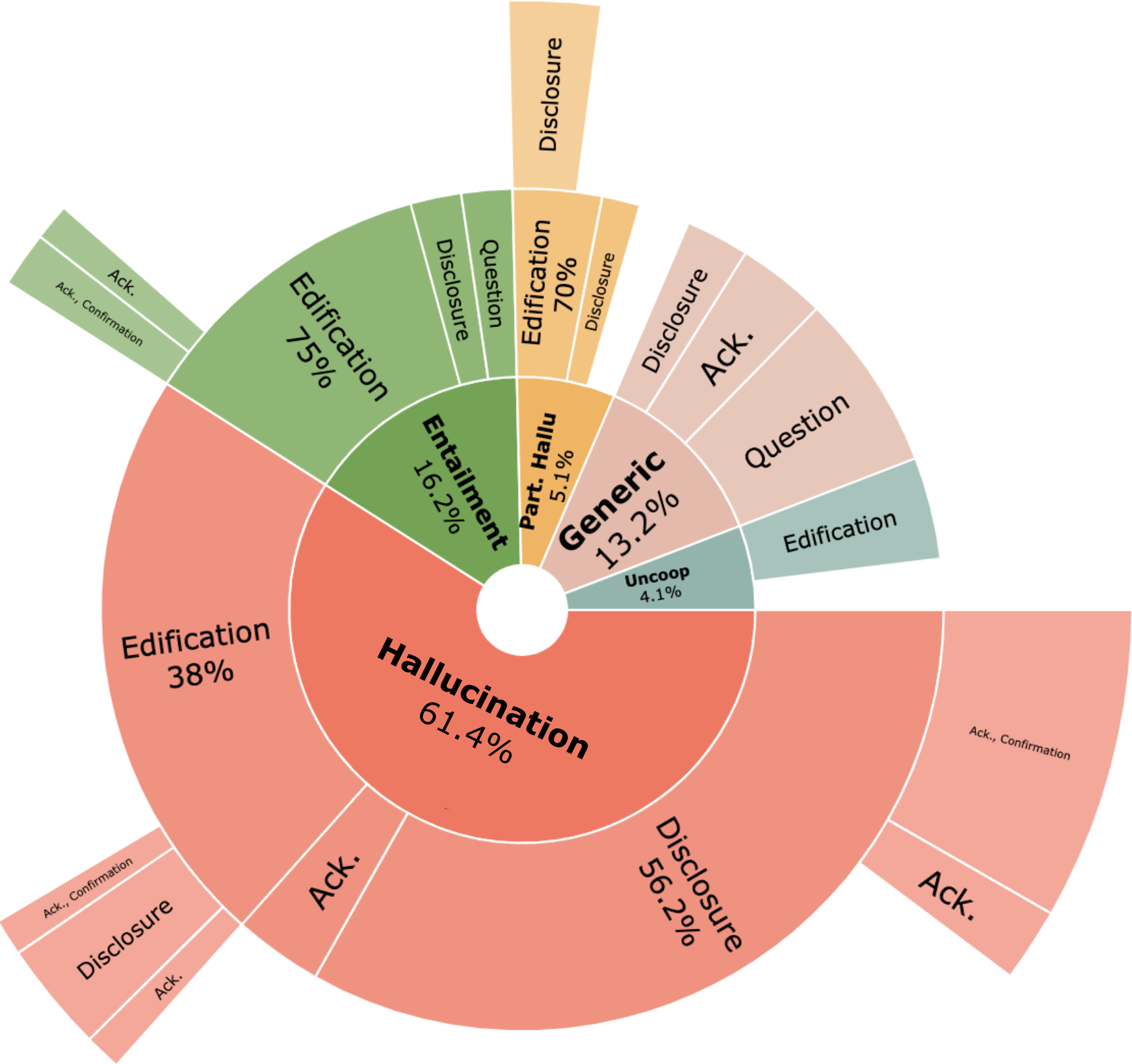}}
    \subfigure[\label{fig:topical_begin}\footnotesize TopicalChat responses]{\includegraphics[width=0.46\linewidth]{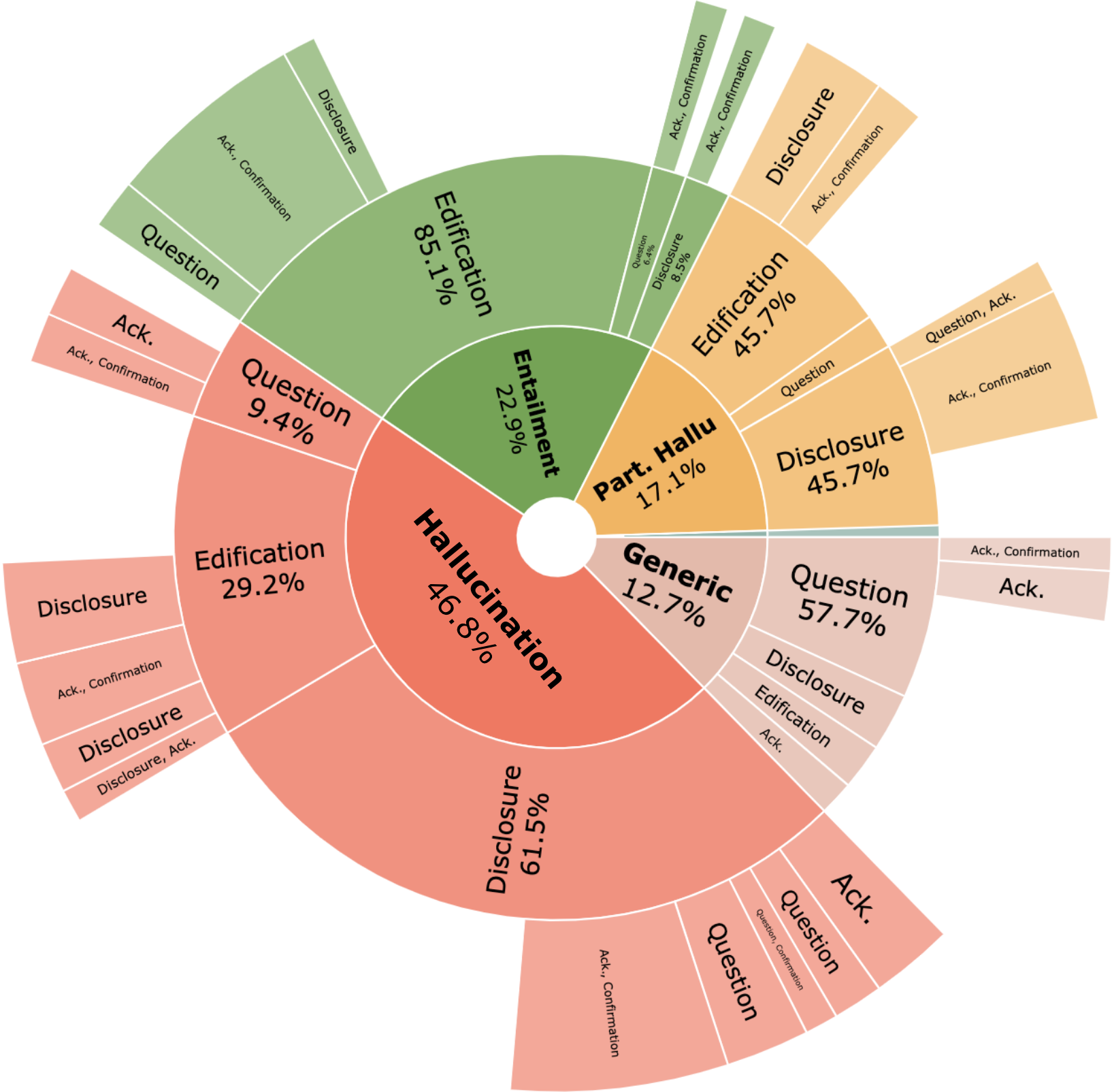}}
    \caption{\small \textsc{BEGIN} and \textsc{VRM} breakdown of gold responses from \textsc{CMU-DoG} and \textsc{TopicalChat}. The inner circle shows the breakdown of BEGIN classes and the outer shows the \textsc{VRM} types in each \textsc{BEGIN} type: Hallucination (red), Entailment (green), Partial Hallucination (yellow), Generic (pink), and Uncooperative (blue).}
    \label{fig:breakdown_gold_cmu_topical}
\end{figure*}

The second round corresponds to a large-scale annotation of 4K randomly sampled train responses using non-expert annotators from AMT. This round is crucial to ensure that the obtained results from the experts are reliable enough to draw conclusions about the quality of the data. As human annotation is expensive, we perform the non-expert annotations only on the \textsc{WoW} benchmark while restricting ourselves to expert annotations on \textsc{CMU-DoG} and \textsc{TopicalChat} data. We choose \textsc{WoW} over the other two datasets as the source knowledge is more amenable to faster annotation (\textsc{TopicalChat}: 300 words > \textsc{CMU-DoG}: 215 words > \textsc{WoW}: 27 words). 
Details about our AMT task design and how we ensure data quality can be found in~\S\ref{amt_ann}. In total, we selected 4 trusted workers to annotate the 4k responses. To compute the inter-annotator agreement, we assign three workers per response in a secondary task, and ask each of them to judge 500 responses. Reported Fleiss' Kappa agreements were 0.75 for \textsc{BEGIN} and 0.61 for \textsc{VRM}. Although substantial, the agreement is lower than the experts' one and this is expected as they have stronger linguistic background. We seek to answer the following questions:

\paragraph{(Q1) How much hallucination exists in the benchmarks?}
Figure \ref{fig:breakdown_gold} shows the breakdown of each \textsc{BEGIN} categoty in \textsc{WoW} and compares expert annotations versus AMT workers.   Surprisingly, \textsc{WoW} is fraught with hallucinations. Expert annotations on 200 responses show that hallucinated responses  are largely mixed with faithful content ($42.3\%$ v.s.  $19.7\%$ fully hallucinated responses), which amounts to $62\%$ hallucinations in total. 
 These results generalize even on larger data; we can see that the portion of hallucinated responses increased to $74.4\%$ when evaluated on 4K samples. Our analysis shows similar trends on the \textsc{CMU-DoG} and \textsc{TopicalChat} benchmarks (Figure~\ref{fig:breakdown_gold_cmu_topical}). \textsc{CMU-DoG} contains $61.4\%$ responses that are purely hallucinated against only $16.2\%$ responses that are fully entailing the source knowledge and \textsc{TopicalChat} has similar results ($63.9\%$ hallucination v.s. $22.9\%$ entailment). Exemplars of hallucinated responses are depicted in~\S\ref{gold_hall_resp}. 
These findings raise the question on the quality of dialogue datasets.

\begin{table}[t!]
    \fontsize{8}{10.5}\selectfont
    \centering
    \begin{tabular}{lp{5.4cm}}
        \toprule
        \textbf{VRM Type} & \textbf{Description}          \\
        \midrule
        \multirow{2}{*}{Disclosure}     
        &  {Reveal the speaker's subjective opinions, personal experience, thoughts and feelings. } 
        \\
  
        \midrule
        \multirow{1}{*}{Edification}       
        & {Concerns information that is objective.} 
        \\
        \midrule
        
        \multirow{2}{*}{Advisement}          
        & {Corresponds to guiding the behaviour of the addressee through:
commands, requests, suggestions, advice, permission, prohibition. } 
        \\
         \midrule
         
        \multirow{2}{*}{Confirmation}         
        & {Compares the speaker’s experience with the other’s by expressing shared ideas or by agreement, disagreement.}  
        \\ 
        \midrule
        
        \multirow{1}{*}{Question}           
        & {Concerns requesting information or guidance. }  
        \\
         \midrule
        
        \multirow{2}{*}{Acknowledge}
        & {Expresses no content, it conveys only receipt of communication from the other's speaker.} 
        \\
        \bottomrule
    \end{tabular}
    \caption{\small{Definitions of the Verbal Response Modes (VRMs)}}
    \vspace*{-3mm}
    \label{fig:vrm_definitions}
\end{table}

 \paragraph{(Q2) What are the hallucination strategies used in human-human data?}
Figure \ref{fig:breakdown_gold}  and Figure \ref{fig:breakdown_gold_cmu_topical} show the VRM breakdown for each BEGIN category in the three benchmarks. 
We make the following observations:
The majority of hallucinations belong to \emph{disclosure} (i.e., subjective information) in all benchmarks (50.9\%, 56.2\% and 61.5\% in \textsc{WoW}, \textsc{CMU-DoG} and \textsc{TopicalChat} respectively). 
Although the strategy of sharing subjective information such as thoughts, opinions and feelings is natural in conversations, it often comes at a cost of ignoring the knowledge snippet in these datasets.
Moreover, \textit{edification} is also a common phenomenon in hallucinated responses, suggesting that humans not only discuss subjective information but also bring extra unsupported facts, either true or false.
Other linguistic modes are also associated with hallucinations such as acknowledging unsupported claims or asking irrelevant questions.
Conversely, entailment responses have high percentage of edification ($>70\%$) with information inferred from the knowledge snippet.

\definecolor{bananamania}{rgb}{0.98, 0.91, 0.71}

\section{Hallucination Amplification in Models}
\label{obj2: hallu_models}
Next, we investigate how much models amplify the hallucination phenomenon at inference time. 
We consider a range of representative models:

\noindent$\bullet$ \textbf{GPT2} \cite{radford2019language, wolf2019transfertransfo} is an autoregressive model which takes as input a concatenation of the knowledge and the history. 

\noindent$\bullet$ \textbf{DoHA} \cite{prabhumoye-etal-2021-focused} builds a BART-based conversational model \cite{lewis-etal-2020-bart} for knowledge-grounding, with a two-view attention mechanism to handle separately the encoded document and the history during generation.

\noindent$\bullet$ \textbf{CTRL} \cite{rashkin-etal-2021-increasing} augments the GPT2 model with control tokens~\cite{keskar2019ctrl} that guide the generation towards less subjective and more entailed content. \\
We fine-tune each model on the  benchmarks and use nucleus sampling \cite{holtzman2019curious} with $p = 0.6$ for decoding (more implementation details are in~\S\ref{app:implementation_details}).
    As seen in \Cref{tab:amplification_model}, CTRL is the best model followed by DoHA based on the hallucination ratio.
    Table~\ref{tab:generated_Resp} in ~\S\ref{app:resp_output} shows a sample of generated responses.
    Similar to the analysis in \S\ref{sec2}, we task the same two linguists to analyze model-generated responses for 200 randomly-selected test samples from each benchmark. 

\begin{table}[t]
\centering
\small
\setlength{\tabcolsep}{2pt}
\scriptsize
\begin{tabular}{ c l | c  c  c  c  c  c  c  c}
 \toprule
 & \multirow{2}{*}{\textbf{Model}} &\multirow{2}{*}{\textbf{R-L$\mathbf{\uparrow}$}} & \multicolumn{3}{c}{\textbf{Hallucination Rate$\downarrow$}}&\multicolumn{3}{c}{\textbf{Entailment Rate$\mathbf{\uparrow}$}}\\
  &&  & {{ \textbf{Full}}} &  {{ \textbf{Partial}}} & {{ \textbf{Overall}}} & {{ \textbf{Entail.}}} &  {{ \textbf{Uncoop.}}} & {{ \textbf{Overall}}} \\
  \midrule
  \multirow{4}{*}{\rotatebox[origin=c]{90}{\scriptsize \texttt{\textbf{WoW}}}} & {{\small \small Gold}}  & \small 36.1  & {\small 19.7} & {\small 42.3}& {\small 62.0} & {\small 24.1}  & \small 8.5  & \small 32.7 \\ 
  &  {{\small \small GPT2}} & {\small{27.0}}  & \small 66.0 & \small 15.2 & \small 81.2  & {\small 11.7}  & \small 3.6 & \small 15.3 \\
  & \multirow{1}{*}{\small \small DoHA} & \small 30.6 & \small 39.6  & \small 28.9   & \small 68.5   & {\small 12.7}&  \small 7.1 & \small 19.8
  \\
  &  {\small CTRL } & {\small 51.3}  &  {\small  31.0} &  {\small 5.0}  & {\small 36.0} & {\small 19.5}& \small 42.0 & \small 61.5
  \\ 
  \midrule
 \multirow{4}{*}{\rotatebox[origin=c]{90}{\scriptsize \texttt{\textbf{CMU-DoG}}}} & {{\small \small Gold}}  &  \small 4.1 &\small 61.4 & {\small 5.1} & {\small 66.5} & {\small 16.2}& {\small 4.1}&  \small 20.3  \\ 
  &  {{\small \small GPT2}} & {\small{4.6}}  & \small 75.5  & \small 6.0 & \small 81.5 & \small   5.5 & \small 5.5 & \small 11.0\\
  & \multirow{1}{*}{\small \small DoHA} &  \small 5.1  & \small 62.5  & \small 10.0   & \small 72.5   & \small   8.5 & \small 5.0 & \small 13.5
  \\ 
    & \multirow{1}{*}{\small \small CTRL} & \small 6.9 &  {\small 62.5} &  {\small 4.5}  & 
    {\small 67.0}  & \small 13.5 & \small 17.0 & \small  30.5
  \\ 
 \midrule
 
\multirow{4}{*}{\rotatebox[origin=c]{90}{\scriptsize \texttt{\textbf{Topical}}}} &
{{\small \small Gold}}  & {\small 1.2} & {\small 46.8} & {\small 17.1}& {\small 63.9} & {\small 22.9} & \small 0.5 & \small 23.4\\ 
  &  {{\small \small GPT2}} & {\small{6.9}}  & \small 70.5 & \small 8.5 & \small 79.0 &  \small 6.5 & \small 5.0 & \small 11.5\\
  & \multirow{1}{*}{\small \small DoHA} & {\small{4.0}}& {\small{53.0}} & \small 25.0 & \small 78.0 &  \small 9.0  & \small  5.0 & \small 14.0 
  \\
  &  {\small CTRL } & {\small 7.9}  &  {\small 48.5} &  {\small 16.7}  & {\small  65.2} & {\small 12.1} & \small 20.7 & \small 32.8
  \\ 
  \bottomrule
\end{tabular}
\caption{\small Amplification of models on the test data from \textsc{WoW} and \textsc{CMU-DoG} and \textsc{TopicalChat}. `Entail.' and `Uncoop.' mean entailment and uncooperative, respectively. R-L measures the ROUGE-L scores between the response and the knowledge.
}
\label{tab:amplification_model}
\vspace{-5mm}
\end{table}

\paragraph{(Q3) Do state-of-the-art conversational models amplify hallucination?}
Table \ref{tab:amplification_model} shows the degree of amplification across different models trained on the three benchmarks. Numbers report the percentage of each class in the data.
Contrasting this with human gold responses, the models not only hallucinate but also amplify the percentage of hallucinations, except \textsc{CTRL} on \textsc{WoW}. 
For example, \textsc{GPT2} amplifies full hallucination by $19.2\%$ in \textsc{WoW}, $15\%$ in \textsc{CMU-DoG} and $15.1\%$ in \textsc{TopicalChat}. 
Conversely, it reduces entailment by $17.4\%$, $9.3\%$ and $11.9\%$ respectively. %
This suggests that  hallucination patterns are easier to learn than entailment.
Among the three, \textsc{CTRL} hallucinates the least at the expense of producing a high number of uncooperative responses. Although these responses are entailing the knowledge, they are not coherent with the history. A closer inspection 
shows that most uncooperative responses are extractive, i.e., they copy big chunks of the evidence without adapting the content to the history or they just output an exact copy of the entire evidence.  This is also reflected in high ROUGE scores between the response and the knowledge, corroborating the extractive nature of \textsc{CTRL} compared to the gold responses. 
This behavior is not surprising as \textsc{CTRL} was optimized to maximize the overlap with the knowledge. Overall, these results demonstrate that hallucination is not only a reflection of training data issues, but also a consequence of the weaknesses of models. 

We hypothesize that there are multiple factors that can contribute to the models' deficiencies: First,  
the exposure bias \cite{DBLP:journals/corr/RanzatoCAZ15} caused by teacher forcing can make hallucination worse as the model may over-rely on previously predicted words which in turn can aggravate error propagation. Second, 
  maximum likelihood estimation can be fragile to noisy data points as it necessitates models to assign high probability mass to all test references, resulting in unstable behavior---a fact observed in machine summarization \cite{kang-hashimoto-2020-improved}. Moreover, we link this issue to the decoding strategies used at test time. We conjecture that models---when conditioned on factual knowledge---often assign the highest probability mass to the correct response and sampling based on other distributions
(e.g. top-k or nucleus) may invite hallucination in the generation process. And lastly, we hypothesise that the behavior of these models is ultimately shaped by the bias learned from internet text during pre-training \cite{nadeem-etal-2021-stereoset}.
We leave investigating the role of each factors to hallucination amplification for future work.

 \paragraph{(Q4) What are the hallucination strategies used by models?}
Surprisingly, different models use different strategies for hallucination.
While DoHA and GPT2 predominantly rely on and amplify \emph{disclosure}, CTRL relies on \emph{edification}.
This is because CTRL is trained explicitly to avoid pronouns (a crucial ingredient for disclosure) and to generate entailed responses.
As a side-effect, it ends up amplifying uncooperative responses (by $33.5\%$, $12.9\%$ and $20.2\%$ in \textsc{WoW} and \textsc{CMU-DoG} as seen in \Cref{tab:amplification_model}).
Full results of all models and datasets are in Figure~\ref{amplification-wow-new}, \ref{amplification-cmu-new} and \ref{amplification-topical-new} in \S\ref{app:vrm_ampl}.

\section{Related Work}
Hallucination in neural language generation has recently attracted the attention of several researchers in many areas including neural machine translation (NMT) \cite{raunak-etal-2021-curious,wang-sennrich-2020-exposure} and summarization \cite{durmus-etal-2020-feqa,kang-hashimoto-2020-improved}.
Hallucinations in knowledge-grounded neural dialogue generation is instead a nascent research problem \cite{mielke2020linguistic,shuster-etal-2021-retrieval-augmentation,dziri-etal-2021-neural, rashkin-etal-2021-increasing}. Most existing works focus on avoiding hallucinations in generated outputs by introducing more robust training approaches. 
 \citet{dziri-etal-2021-neural} propose a model that uses facts supplied by a knowledge graph to reduce entity-based hallucinations in generated responses. \citet{rashkin-etal-2021-increasing}
add control tokens at training time to control generation towards more objective sentences and faithful sentences. 
Closest to our work are \citet{dziri2021evaluating} and \citet{rashkin2021measuring} who introduce frameworks for quantifying attribution in dialogue systems, whereas we conduct a much finer-grained manual analysis on multiple benchmarks and models.

\section{Conclusion}
Our investigations demonstrate empirically that hallucination is a prevalent issue in both dialog benchmarks and models.
Our analysis on three widely used benchmarks reveals that they are rife with hallucinations, and the most common strategies people use are \emph{disclosure} and \emph{edification}.
Moreover, we show that conversational models trained on these benchmarks not only hallucinate but also amplify hallucinations, even the models that were designed to alleviate this issue.
This calls for a clean high-quality data release and careful design of trustworthy conversational systems. Before then, we strongly advocate practitioners to look at samples of any dataset---in order to uncover actionable insights---prior to their use or public release.

\section*{Acknowledgements}
We are grateful to the anonymous reviewers for helpful comments. 
This research is supported by the Mila-IBM grant and the  Alberta Machine Intelligence Institute Fellow Program.
We also acknowledge the support of the NSERC Discovery grant and the Facebook CIFAR AI Chair.

\newpage
\section*{Impact Statement \& Ethics}
\paragraph{Annotation Risks}
The benchmarks we audit were collected through AMT and thus may contain some disturbing examples including racist or even expletive phrases.
Annotators were also asked to judge the outputs of several state-of-the-art conversational systems which may be in turn toxic and insensitive. We acknowledge the psychological distress that this may present to workers \cite{arditte2016importance}. Therefore, we alert workers by adding the following warning in italic text in each HIT: \emph{If this HIT causes you emotional distress or elicit feelings of trauma, please feel free to skip it. }

\paragraph{Deployment Risks} Our analytical study reveals that a large portion of standard knowledge-grounded dialogue benchmarks is hallucinated, leading us to reflect on the potential harm of low-quality data releases for conversational models. In recent years, the conversational AI market has seen a proliferation of a variety of applications---which are powered by large pre-trained LMs---that span across a broad range of domains, such as customer support, education, e-commerce, health, entertainment,  etc \cite{vakulenko2021large}. 
Ensuring that these systems are trustworthy is key to deploy systems safely at a large scale in real-world application, especially in high-stake domains \cite{sambasivan2021everyone}. However, even if we come up with a model that is robust enough against hallucination, it will be ultimately bounded by the data quality. We argue that fixing the models or the data to enforce faithfulness is a highly non-trivial task without an in-depth understanding of the various sources of hallucination. Our work thus represents the first effort to gain such an understanding and to inform the community about the unreliability of the existing benchmarks and models. As result, we believe it is important to raise these insights to the broader community.

\bibliography{anthology,custom}

\begin{thebibliography}{34}
\expandafter\ifx\csname natexlab\endcsname\relax\def\natexlab#1{#1}\fi

\bibitem[{Arditte et~al.(2016)Arditte, {\c{C}}ek, Shaw, and
  Timpano}]{arditte2016importance}
Kimberly~A Arditte, Demet {\c{C}}ek, Ashley~M Shaw, and Kiara~R Timpano. 2016.
\newblock The importance of assessing clinical phenomena in mechanical turk
  research.
\newblock \emph{Psychological assessment}, 28(6):684.

\bibitem[{Bender et~al.(2021)Bender, Gebru, McMillan-Major, and
  Shmitchell}]{bender2021dangers}
Emily~M Bender, Timnit Gebru, Angelina McMillan-Major, and Shmargaret
  Shmitchell. 2021.
\newblock On the dangers of stochastic parrots: Can language models be too big?
\newblock In \emph{Proceedings of the 2021 ACM Conference on Fairness,
  Accountability, and Transparency}, pages 610--623.

\bibitem[{Brown et~al.(2020)Brown, Mann, Ryder, Subbiah, Kaplan, Dhariwal,
  Neelakantan, Shyam, Sastry, Askell, Agarwal, Herbert-Voss, Krueger, Henighan,
  Child, Ramesh, Ziegler, Wu, Winter, Hesse, Chen, Sigler, Litwin, Gray, Chess,
  Clark, Berner, McCandlish, Radford, Sutskever, and
  Amodei}]{NEURIPS2020_1457c0d6}
Tom Brown, Benjamin Mann, Nick Ryder, Melanie Subbiah, Jared~D Kaplan, Prafulla
  Dhariwal, Arvind Neelakantan, Pranav Shyam, Girish Sastry, Amanda Askell,
  Sandhini Agarwal, Ariel Herbert-Voss, Gretchen Krueger, Tom Henighan, Rewon
  Child, Aditya Ramesh, Daniel Ziegler, Jeffrey Wu, Clemens Winter, Chris
  Hesse, Mark Chen, Eric Sigler, Mateusz Litwin, Scott Gray, Benjamin Chess,
  Jack Clark, Christopher Berner, Sam McCandlish, Alec Radford, Ilya Sutskever,
  and Dario Amodei. 2020.
\newblock \href
  {https://proceedings.neurips.cc/paper/2020/file/1457c0d6bfcb4967418bfb8ac142f64a-Paper.pdf}
  {Language models are few-shot learners}.
\newblock In \emph{Advances in Neural Information Processing Systems},
  volume~33, pages 1877--1901. Curran Associates, Inc.

\bibitem[{Bunt et~al.(2020)Bunt, Petukhova, Gilmartin, Pelachaud, Fang, Keizer,
  and Pr{\'e}vot}]{bunt-etal-2020-iso}
Harry Bunt, Volha Petukhova, Emer Gilmartin, Catherine Pelachaud, Alex Fang,
  Simon Keizer, and Laurent Pr{\'e}vot. 2020.
\newblock \href {https://aclanthology.org/2020.lrec-1.69} {The {ISO} standard
  for dialogue act annotation, second edition}.
\newblock In \emph{Proceedings of the 12th Language Resources and Evaluation
  Conference}, pages 549--558, Marseille, France. European Language Resources
  Association.

\bibitem[{Dinan et~al.(2018)Dinan, Roller, Shuster, Fan, Auli, and
  Weston}]{dinan2018wizard}
Emily Dinan, Stephen Roller, Kurt Shuster, Angela Fan, Michael Auli, and Jason
  Weston. 2018.
\newblock Wizard of wikipedia: Knowledge-powered conversational agents.
\newblock In \emph{International Conference on Learning Representations}.

\bibitem[{Durmus et~al.(2020)Durmus, He, and Diab}]{durmus-etal-2020-feqa}
Esin Durmus, He~He, and Mona Diab. 2020.
\newblock \href {https://doi.org/10.18653/v1/2020.acl-main.454} {{FEQA}: A
  question answering evaluation framework for faithfulness assessment in
  abstractive summarization}.
\newblock In \emph{Proceedings of the 58th Annual Meeting of the Association
  for Computational Linguistics}, pages 5055--5070, Online. Association for
  Computational Linguistics.

\bibitem[{Dziri et~al.(2021{\natexlab{a}})Dziri, Madotto, Za{\"\i}ane, and
  Bose}]{dziri-etal-2021-neural}
Nouha Dziri, Andrea Madotto, Osmar Za{\"\i}ane, and Avishek~Joey Bose.
  2021{\natexlab{a}}.
\newblock \href {https://aclanthology.org/2021.emnlp-main.168} {Neural path
  hunter: Reducing hallucination in dialogue systems via path grounding}.
\newblock In \emph{Proceedings of the 2021 Conference on Empirical Methods in
  Natural Language Processing}, pages 2197--2214, Online and Punta Cana,
  Dominican Republic. Association for Computational Linguistics.

\bibitem[{Dziri et~al.(2021{\natexlab{b}})Dziri, Rashkin, Linzen, and
  Reitter}]{dziri2021evaluating}
Nouha Dziri, Hannah Rashkin, Tal Linzen, and David Reitter. 2021{\natexlab{b}}.
\newblock Evaluating groundedness in dialogue systems: The begin benchmark.
\newblock \emph{arXiv preprint arXiv:2105.00071}.

\bibitem[{Gopalakrishnan et~al.(2019)Gopalakrishnan, Hedayatnia, Chen,
  Gottardi, Kwatra, Venkatesh, Gabriel, and Hakkani-Tür}]{Gopalakrishnan2019}
Karthik Gopalakrishnan, Behnam Hedayatnia, Qinlang Chen, Anna Gottardi, Sanjeev
  Kwatra, Anu Venkatesh, Raefer Gabriel, and Dilek Hakkani-Tür. 2019.
\newblock \href {https://doi.org/10.21437/Interspeech.2019-3079}
  {{Topical-Chat: Towards Knowledge-Grounded Open-Domain Conversations}}.
\newblock In \emph{Proc. Interspeech 2019}, pages 1891--1895.

\bibitem[{Grice(1989)}]{grice1989studies}
Paul Grice. 1989.
\newblock \emph{Studies in the Way of Words}.
\newblock Harvard University Press.

\bibitem[{Holtzman et~al.(2019)Holtzman, Buys, Du, Forbes, and
  Choi}]{holtzman2019curious}
Ari Holtzman, Jan Buys, Li~Du, Maxwell Forbes, and Yejin Choi. 2019.
\newblock The curious case of neural text degeneration.
\newblock In \emph{International Conference on Learning Representations}.

\bibitem[{Kang and Hashimoto(2020)}]{kang-hashimoto-2020-improved}
Daniel Kang and Tatsunori~B. Hashimoto. 2020.
\newblock \href {https://doi.org/10.18653/v1/2020.acl-main.66} {Improved
  natural language generation via loss truncation}.
\newblock In \emph{Proceedings of the 58th Annual Meeting of the Association
  for Computational Linguistics}, pages 718--731, Online. Association for
  Computational Linguistics.

\bibitem[{Keskar et~al.(2019)Keskar, McCann, Varshney, Xiong, and
  Socher}]{keskar2019ctrl}
Nitish~Shirish Keskar, Bryan McCann, Lav~R Varshney, Caiming Xiong, and Richard
  Socher. 2019.
\newblock Ctrl: A conditional transformer language model for controllable
  generation.
\newblock \emph{arXiv preprint arXiv:1909.05858}.

\bibitem[{Kingma and Ba(2015)}]{KingmaB14}
Diederik~P. Kingma and Jimmy Ba. 2015.
\newblock \href {http://arxiv.org/abs/1412.6980} {Adam: A method for stochastic
  optimization}.
\newblock In \emph{ICLR (Poster)}.

\bibitem[{Lewis et~al.(2020)Lewis, Liu, Goyal, Ghazvininejad, Mohamed, Levy,
  Stoyanov, and Zettlemoyer}]{lewis-etal-2020-bart}
Mike Lewis, Yinhan Liu, Naman Goyal, Marjan Ghazvininejad, Abdelrahman Mohamed,
  Omer Levy, Veselin Stoyanov, and Luke Zettlemoyer. 2020.
\newblock \href {https://doi.org/10.18653/v1/2020.acl-main.703} {{BART}:
  Denoising sequence-to-sequence pre-training for natural language generation,
  translation, and comprehension}.
\newblock In \emph{Proceedings of the 58th Annual Meeting of the Association
  for Computational Linguistics}, pages 7871--7880, Online. Association for
  Computational Linguistics.

\bibitem[{Mielke et~al.(2020)Mielke, Szlam, Boureau, and
  Dinan}]{mielke2020linguistic}
Sabrina~J Mielke, Arthur Szlam, Y-Lan Boureau, and Emily Dinan. 2020.
\newblock Linguistic calibration through metacognition: aligning dialogue agent
  responses with expected correctness.
\newblock \emph{arXiv preprint arXiv:2012.14983}.

\bibitem[{Moon et~al.(2019)Moon, Shah, Kumar, and
  Subba}]{moon-etal-2019-opendialkg}
Seungwhan Moon, Pararth Shah, Anuj Kumar, and Rajen Subba. 2019.
\newblock \href {https://doi.org/10.18653/v1/P19-1081} {{O}pen{D}ial{KG}:
  Explainable conversational reasoning with attention-based walks over
  knowledge graphs}.
\newblock In \emph{Proceedings of the 57th Annual Meeting of the Association
  for Computational Linguistics}, pages 845--854, Florence, Italy. Association
  for Computational Linguistics.

\bibitem[{Nadeem et~al.(2021)Nadeem, Bethke, and
  Reddy}]{nadeem-etal-2021-stereoset}
Moin Nadeem, Anna Bethke, and Siva Reddy. 2021.
\newblock \href {https://doi.org/10.18653/v1/2021.acl-long.416} {{S}tereo{S}et:
  Measuring stereotypical bias in pretrained language models}.
\newblock In \emph{Proceedings of the 59th Annual Meeting of the Association
  for Computational Linguistics and the 11th International Joint Conference on
  Natural Language Processing (Volume 1: Long Papers)}, pages 5356--5371,
  Online. Association for Computational Linguistics.

\bibitem[{Prabhumoye et~al.(2021)Prabhumoye, Hashimoto, Zhou, Black, and
  Salakhutdinov}]{prabhumoye-etal-2021-focused}
Shrimai Prabhumoye, Kazuma Hashimoto, Yingbo Zhou, Alan~W Black, and Ruslan
  Salakhutdinov. 2021.
\newblock \href {https://doi.org/10.18653/v1/2021.naacl-main.338} {Focused
  attention improves document-grounded generation}.
\newblock In \emph{Proceedings of the 2021 Conference of the North American
  Chapter of the Association for Computational Linguistics: Human Language
  Technologies}, pages 4274--4287, Online. Association for Computational
  Linguistics.

\bibitem[{Radford et~al.(2019)Radford, Wu, Child, Luan, Amodei, Sutskever
  et~al.}]{radford2019language}
Alec Radford, Jeffrey Wu, Rewon Child, David Luan, Dario Amodei, Ilya
  Sutskever, et~al. 2019.
\newblock Language models are unsupervised multitask learners.
\newblock \emph{OpenAI blog}, 1(8):9.

\bibitem[{Raffel et~al.(2020)Raffel, Shazeer, Roberts, Lee, Narang, Matena,
  Zhou, Li, and Liu}]{2020t5}
Colin Raffel, Noam Shazeer, Adam Roberts, Katherine Lee, Sharan Narang, Michael
  Matena, Yanqi Zhou, Wei Li, and Peter~J. Liu. 2020.
\newblock \href {http://jmlr.org/papers/v21/20-074.html} {Exploring the limits
  of transfer learning with a unified text-to-text transformer}.
\newblock \emph{Journal of Machine Learning Research}, 21(140):1--67.

\bibitem[{Ranzato et~al.(2016)Ranzato, Chopra, Auli, and
  Zaremba}]{DBLP:journals/corr/RanzatoCAZ15}
Marc'Aurelio Ranzato, Sumit Chopra, Michael Auli, and Wojciech Zaremba. 2016.
\newblock \href {http://arxiv.org/abs/1511.06732} {Sequence level training with
  recurrent neural networks}.
\newblock In \emph{4th International Conference on Learning Representations,
  {ICLR} 2016, San Juan, Puerto Rico, May 2-4, 2016, Conference Track
  Proceedings}.

\bibitem[{Rashkin et~al.(2021{\natexlab{a}})Rashkin, Nikolaev, Lamm, Collins,
  Das, Petrov, Tomar, Turc, and Reitter}]{rashkin2021measuring}
Hannah Rashkin, Vitaly Nikolaev, Matthew Lamm, Michael Collins, Dipanjan Das,
  Slav Petrov, Gaurav~Singh Tomar, Iulia Turc, and David Reitter.
  2021{\natexlab{a}}.
\newblock Measuring attribution in natural language generation models.
\newblock \emph{arXiv preprint arXiv:2112.12870}.

\bibitem[{Rashkin et~al.(2021{\natexlab{b}})Rashkin, Reitter, Tomar, and
  Das}]{rashkin-etal-2021-increasing}
Hannah Rashkin, David Reitter, Gaurav~Singh Tomar, and Dipanjan Das.
  2021{\natexlab{b}}.
\newblock \href {https://doi.org/10.18653/v1/2021.acl-long.58} {Increasing
  faithfulness in knowledge-grounded dialogue with controllable features}.
\newblock In \emph{Proceedings of the 59th Annual Meeting of the Association
  for Computational Linguistics and the 11th International Joint Conference on
  Natural Language Processing (Volume 1: Long Papers)}, pages 704--718, Online.
  Association for Computational Linguistics.

\bibitem[{Raunak et~al.(2021)Raunak, Menezes, and
  Junczys-Dowmunt}]{raunak-etal-2021-curious}
Vikas Raunak, Arul Menezes, and Marcin Junczys-Dowmunt. 2021.
\newblock \href {https://doi.org/10.18653/v1/2021.naacl-main.92} {The curious
  case of hallucinations in neural machine translation}.
\newblock In \emph{Proceedings of the 2021 Conference of the North American
  Chapter of the Association for Computational Linguistics: Human Language
  Technologies}, pages 1172--1183, Online. Association for Computational
  Linguistics.

\bibitem[{Sambasivan et~al.(2021)Sambasivan, Kapania, Highfill, Akrong,
  Paritosh, and Aroyo}]{sambasivan2021everyone}
Nithya Sambasivan, Shivani Kapania, Hannah Highfill, Diana Akrong, Praveen
  Paritosh, and Lora~M Aroyo. 2021.
\newblock “everyone wants to do the model work, not the data work”: Data
  cascades in high-stakes ai.
\newblock In \emph{proceedings of the 2021 CHI Conference on Human Factors in
  Computing Systems}, pages 1--15.

\bibitem[{Shuster et~al.(2021)Shuster, Poff, Chen, Kiela, and
  Weston}]{shuster-etal-2021-retrieval-augmentation}
Kurt Shuster, Spencer Poff, Moya Chen, Douwe Kiela, and Jason Weston. 2021.
\newblock \href {https://aclanthology.org/2021.findings-emnlp.320} {Retrieval
  augmentation reduces hallucination in conversation}.
\newblock In \emph{Findings of the Association for Computational Linguistics:
  EMNLP 2021}, pages 3784--3803, Punta Cana, Dominican Republic. Association
  for Computational Linguistics.

\bibitem[{Srivastava et~al.(2014)Srivastava, Hinton, Krizhevsky, Sutskever, and
  Salakhutdinov}]{srivastava2014dropout}
Nitish Srivastava, Geoffrey Hinton, Alex Krizhevsky, Ilya Sutskever, and Ruslan
  Salakhutdinov. 2014.
\newblock Dropout: a simple way to prevent neural networks from overfitting.
\newblock \emph{The journal of machine learning research}, 15(1):1929--1958.

\bibitem[{Stiles(1992)}]{stiles1992describing}
William~B Stiles. 1992.
\newblock \emph{Describing talk: A taxonomy of verbal response modes}.
\newblock Sage Publications.

\bibitem[{Vakulenko et~al.(2021)Vakulenko, Kanoulas, and
  de~Rijke}]{vakulenko2021large}
Svitlana Vakulenko, Evangelos Kanoulas, and Maarten de~Rijke. 2021.
\newblock A large-scale analysis of mixed initiative in information-seeking
  dialogues for conversational search.
\newblock \emph{arXiv preprint arXiv:2104.07096}.

\bibitem[{Wang and Sennrich(2020)}]{wang-sennrich-2020-exposure}
Chaojun Wang and Rico Sennrich. 2020.
\newblock \href {https://doi.org/10.18653/v1/2020.acl-main.326} {On exposure
  bias, hallucination and domain shift in neural machine translation}.
\newblock In \emph{Proceedings of the 58th Annual Meeting of the Association
  for Computational Linguistics}, pages 3544--3552, Online. Association for
  Computational Linguistics.

\bibitem[{Wolf et~al.(2020)Wolf, Debut, Sanh, Chaumond, Delangue, Moi, Cistac,
  Rault, Louf, Funtowicz, Davison, Shleifer, von Platen, Ma, Jernite, Plu, Xu,
  Le~Scao, Gugger, Drame, Lhoest, and Rush}]{wolf-etal-2020-transformers}
Thomas Wolf, Lysandre Debut, Victor Sanh, Julien Chaumond, Clement Delangue,
  Anthony Moi, Pierric Cistac, Tim Rault, Remi Louf, Morgan Funtowicz, Joe
  Davison, Sam Shleifer, Patrick von Platen, Clara Ma, Yacine Jernite, Julien
  Plu, Canwen Xu, Teven Le~Scao, Sylvain Gugger, Mariama Drame, Quentin Lhoest,
  and Alexander Rush. 2020.
\newblock \href {https://doi.org/10.18653/v1/2020.emnlp-demos.6} {Transformers:
  State-of-the-art natural language processing}.
\newblock In \emph{Proceedings of the 2020 Conference on Empirical Methods in
  Natural Language Processing: System Demonstrations}, pages 38--45, Online.
  Association for Computational Linguistics.

\bibitem[{Wolf et~al.(2019)Wolf, Sanh, Chaumond, and
  Delangue}]{wolf2019transfertransfo}
Thomas Wolf, Victor Sanh, Julien Chaumond, and Clement Delangue. 2019.
\newblock Transfertransfo: A transfer learning approach for neural network
  based conversational agents.
\newblock \emph{arXiv preprint arXiv:1901.08149}.

\bibitem[{Zhou et~al.(2018)Zhou, Prabhumoye, and
  Black}]{zhou-etal-2018-dataset}
Kangyan Zhou, Shrimai Prabhumoye, and Alan~W Black. 2018.
\newblock \href {https://doi.org/10.18653/v1/D18-1076} {A dataset for document
  grounded conversations}.
\newblock In \emph{Proceedings of the 2018 Conference on Empirical Methods in
  Natural Language Processing}, pages 708--713, Brussels, Belgium. Association
  for Computational Linguistics.

\end{thebibliography}
\bibliographystyle{acl_natbib}

\clearpage
\appendix

\section{Datasets}
\label{appendix:dataset}
We conduct our analysis on the following datasets:
\paragraph{Wizard of Wikipedia:} composed of dialogues between a ``wizard'' and an ``apprentice'', where the goal of the wizard is to communicate information about a particular topic and the apprentice is tasked to seek information about that topic. At each turn, the wizard is presented with a knowledge snippet from Wikipedia and asked to form an utterance. We
filter data points in which the wizard did not explicitly select a passage as knowledge for the response. In total, the dataset is comprised of 82722 grounded-responses in train,  8800 valid and 8690 test.

\paragraph{CMU-DoG:} All conversations focus only on the movie domain. Each response is grounded on a section from Wikipedia. Workers are asked to either persuade the other speakers to watch the movie using information from the Wikipedia section or to discuss the content of the document with them. In total, there are 78136 grounded responses in train, 13800 in valid and 13796 in test.
  
\paragraph{TopicalChat:} Contrary to \textsc{CMU-DoG},  \textsc{TopicalChat} conversations are about a variety of topics. Workers are provided relevant facts from Reddit, Wikipedia and news articles. The collection process corresponds to two scenarios: symmetric and asymmetric. In the symmetric scenario, workers  have access to the same source knowledge and in the asymmetric scenario, they have access to different sources. In total, the dataset has 292215 grounded responses in train, 23601 in valid and 23623 in test.

\section{Implementation Details}
\label{app:implementation_details}

\paragraph{GPT2:} This model was implemented using the  Pytorch Huggingface Transformers library \cite{wolf-etal-2020-transformers}  and  the Pytorch-lightning library\footnote{\url{https://github.com/PyTorchLightning/pytorch-lightning}}. To train the models, we use the Adam optimizer \cite{KingmaB14} with Dropout \cite{srivastava2014dropout} on a batch size of $32$ with a learning rate of $6.25 \times 10^{-5}$ that is linearly decayed. The maximum dialogue history length is set to $3$ utterances. The model early-stops at epoch \{7, 8, 8\} respectively for \textsc{WoW}, \textsc{CMU-DoG} and \textsc{TopicalChat}.  The average runtime  is \{1.5, 3, 3\}  hours for \textsc{WoW}, \textsc{CMU-DoG} and \textsc{TopicalChat} respectively. 

\paragraph{DoHA:} We use the pre-trained model on CMU-DoG that is publicly available\footnote{\url{https://bit.ly/3bBup2M}}. However, since no models trained on \textsc{WoW} and \textsc{TopicalChat} have been released, we follow closely the training procedure described in \citet{prabhumoye-etal-2021-focused} and we train two models.  The average runtime of these models is \{5, 10\} hours for \textsc{WoW} and \textsc{TopicalChat} respectively.

\paragraph{CTRL:} We implement the model ourselves since the code and the model were not released by the authors. We follow training details in \citet{rashkin-etal-2021-increasing} and implement this model using the  Pytorch Huggingface Transformers library and  the Pytorch-lightning library. Additionally, we had multiple discussions with the authors to make sure that our implementation is accurate. 

We save the best model based on the validation set, for all datasets. Training for all models is done on an Nvidia V100 GPU 32GB and for inference, we use nucleus sampling with p=0.6.

\section{Definition of VRM}
\label{vrm}
Table \ref{fig:vrm_definitions_complete} contains VRM definitions with examples.
\begin{table*}[t!]
    \fontsize{8}{10.5}\selectfont
    \centering
    \begin{tabular}{lp{7.4cm}p{5.9cm}}
        \toprule
        \textbf{VRM Type} & \textbf{Description} & \textbf{Example}         \\
        \midrule
        \multirow{2}{*}{Disclosure}     
        &  {Reveal the speaker's subjective opinions, personal experience, thoughts, feelings, wishes, and intentions. } 
        & {\textit{``I think science fiction is an amazing genre. Future science, technology they're all interesting."}} \\
  
        \midrule
        \multirow{1}{*}{Edification}       
        & {Concerns information that is, in principle, objective.} 
        & \multirow{1}{5.9cm}{\textit{``Recycling includes items like metal and plastic."}} \\
        \midrule
        
        \multirow{2}{*}{Advisement}          
        & {Corresponds to guiding the behaviour of the addressee through:
commands, requests, suggestions, advice, permission, prohibition. } 
        &\multirow{2}{5.9cm}{\textit{``You should be patient and persistent to succeed."}} \\
         \midrule
         
        \multirow{2}{*}{Confirmation}         
        & {Compares the speaker’s experience with the other’s by expressing shared ideas/memories/beliefs, or by agreement/disagreement}  
        & \multirow{2}{5.6cm}{\textit{``I agree that love encompasses a variety of different emotional and mental states."}}
        \\ 
        \midrule
        
        \multirow{1}{*}{Question}           
        & {Concerns requesting information or guidance. }  
        & \multirow{1}{5.9cm}{\textit{``What is your favorite song?"}} \\
         \midrule
        
        \multirow{2}{*}{Acknowledge}
        & {Expresses no content, it conveys only receipt of communication from the other's speaker.} 
        & \multirow{2}{5.9cm}{\textit{“Mmm. OK,...”, “Yeah, ...”, “Hello, ...”}} \\
        \bottomrule
    \end{tabular}
    \caption{\small{The definitions of the VRM types with examples.}}
    \vspace*{-3mm}
    \label{fig:vrm_definitions_complete}
\end{table*}

\begin{table}[t]
    \centering
    \begin{tabular}{c c c}
    \hline
         &  BEGIN & VRM \\
         \midrule
    \textsc{CMU-DoG}     & 0.85 & 0.78 \\
     \textsc{TopicalChat} & 0.83 & 0.72\\
    \hline
    \end{tabular}
    \caption{Fleiss Kappa Scores on 200 train Human-Human responses from the \textsc{CMU-DoG} and \textsc{TopicalChat} benchmarks. }
    \label{tab:kappa}
\end{table}

\begin{figure*}[t]
    \centering
    \includegraphics[width=\linewidth]{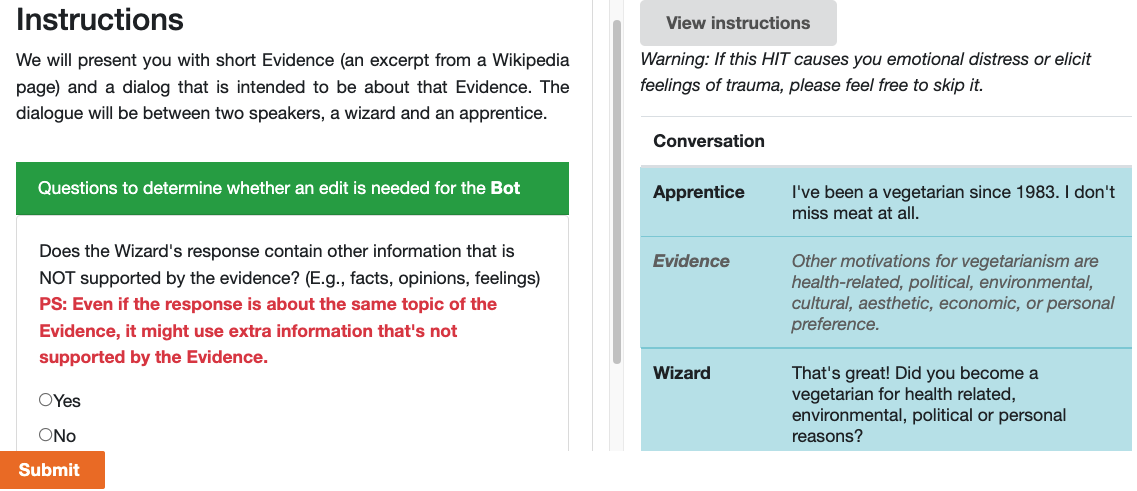}
    \caption{\small{AMT Annotation interfaces for determining BEGIN and VRM categories (1).}}
    \label{fig:amt1}
\end{figure*}

\begin{figure*}[t]
    \centering
    \includegraphics[width=\linewidth]{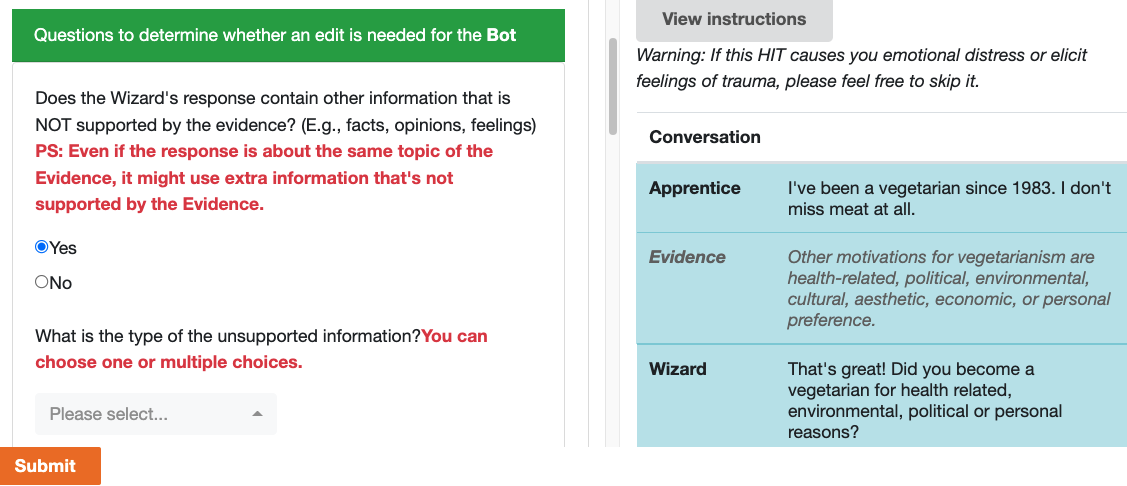}
    \caption{AMT Annotation interfaces for determining BEGIN and VRM categories (2).}
    \label{fig:amt2}
    \vspace{-15pt}
\end{figure*}

\label{app:definitions}

\begin{table*}[ht]
\footnotesize
    \centering
    \begin{tabular}{ L{2cm} L{2cm} l L{9.3cm}}
    \toprule
    \textbf{BEGIN} & \textbf{VRM} & \textbf{Example} \\
    \midrule
    Entailment  & \textcolor{red}{Disclosure}  &
        Knowledge: & A dragon is a legendary creature, typically scaled or fire-spewing and with serpentine, reptilian or avian traits, that features in the myths of many cultures around world.\\
        &&
        History: &Dragons are so fascinating, I wonder where they originated from. \\
        &&
        \textbf{Response}: & I'm not sure, but \textcolor{red}{I know that it is a legendary creature featured in myths of many cultures around the world}! 
        \\ \midrule
        Hallucination & \textcolor{blue}{Edification} \textcolor{red}{Disclosure}  &
        Knowledge: & The central premise for these stories oftentimes involves changing history, either intentionally or by accident, and the ways by which altering the past changes the future and creates an altered present or future for the time traveler when they return home. \\
        && History: & One of my favorite forms of science fiction is anything related to time travel! I find it fascinating.\\
        && \textbf{Response}: & \textcolor{blue}{It's not quite sci-fi}, \textcolor{red}{but my favorite version of time travel is in Harry Potter and the Prisoner of Azkaban}. \textcolor{blue}{Breaks zero logical rules}. 
                            \\ \midrule
        
         Partial hallucination  & \colorbox{babypink}{Confirmation}   &
      Knowledge: &
        Love encompasses a variety of different emotional and mental states, typically strongly and positively experienced, ranging from the deepest interpersonal affection to the simplest pleasure.
        \\
        &&
        History: & Romance can be small acts, like making breakfast for your significant other. But it's portrayed as grand gestures which are unattainable.
        \\
        &&
        \textbf{Response}: & \colorbox{babypink}{I agree it has been portrayed as impossible actions}. Love consists of a variety of emotional and mental states.
        \\ \midrule
        Generic &Question & 
        Knowledge: & Hard rock is a loosely defined subgenre of rock music that began in the mid-1960s, with the garage, psychedelic and blues rock movements.\\
        && History: & yea i have heard of the American hard rock band Van Halen.\\
        && \textbf{Response}: &  Are you a fan of hard rock?.\\
         \midrule
        
           Uncooperative &\textcolor{blue}{Edification}& 
       Knowledge: & Marketing is used to create, keep and satisfy the customer.\\
        && History: & Well, I wish I could. I think my training was a little off. I can't find a job and thinking about going back to school to be a counselor. What do you do?\\
        && \textbf{Response}: &  \textcolor{blue}{Marketing is used to create a customer and then retain the customer and keep them happy}. \\

        \bottomrule
    \end{tabular}
    \caption{Examples from Wizard of Wikipedia \cite{dinan2018wizard} showing the BEGIN breakdown and different VRM linguistic phenomena for each response.}
    \label{tab:benchmark:wow_begin_dist}
\end{table*}

\begin{table*}[ht]
\footnotesize
    \centering
    \begin{tabular}{l L{7cm} L{6.8cm}}
    \toprule
    \textbf{Data} & \textbf{Knowledge \& History} & \textbf{Responses} \\
    \midrule
        \textbf{WoW}  &
        \multirow{6}{7cm}{\textbf{Knowledge}: Cinematography (also called "Direction of Photography") is the science or art of motion-picture photography by recording light or other electromagnetic radiation, either electronically by means of an image sensor, or chemically by means of a light-sensitive material such as film stock. \\ \textbf{History}: Hi buddy, What you think about cinematography?} & \textbf{GPT2}: Cinematography is another title I'm currently in school for. I've been doing the art of photography for over 15 years now. (\underline{Hallucination}) \\
        & & \textbf{DoHA}: I do enjoy the science and art of motion-picture photography. (\underline{Partial Hallucination})\\
        & & \textbf{CTRL}: You're right. Cinematography is an art of motion -picture photography by recording light or other electromagnetic radiation, either electronically by means of an image sensor, or chemically by means of a light-sensitive material such as film stock. What about you ? (\underline{Entailment})\\
        \midrule
        \textbf{CMU-DoG} &
        \multirow{6}{7cm}{\textbf{Knowledge}: Mean Girls is a 2004 American teen comedy film directed by Mark Waters and written by Tina Fey. The film is partially based on Rosalind Wiseman's 2002 non-fiction self-help book Queen Bees and Wannabes, which describes female high school social cliques and the damaging effects they can have on girls. Genre: Comedy. MovieName: Mean Girls rating: Rotten Tomatoes: 84 \% and average: 6.9/10.
        \\ \textbf{History}: I think Rachel McAdams had an even better role as Regina George however!}
        &\textbf{GPT2}: Mean Girls was a classic. (\underline{Hallucination}) 
        \\
        &&
        \textbf{DoHA}: Yes I agree! she was great and she was better than Mark Waters! (\underline{Hallucination})\\
        & & \textbf{CTRL}: I see. Mean Girls is a comedy movie that was directed by Mark Waters in 2004.  (\underline{Entailment}) \\ \\ \\ \\ \\ \\  \\
        \bottomrule
    \end{tabular}
    \caption{Generated responses from different models based on Wizard of Wikipedia \cite{dinan2018wizard} and \textsc{CMU-DoG} \cite{zhou-etal-2018-dataset} test samples. }
    \label{tab:generated_Resp}
\end{table*}

\begin{table*}[ht]
\footnotesize
    \centering
    \begin{tabular}{l L{7cm} L{6.8cm}}
    \toprule
    \textbf{Data} & \textbf{Knowledge \& History} & \textbf{Hallucinated Gold Response} \\
    \midrule
        \textbf{WoW}  &
        \multirow{6}{7cm}{\textbf{Knowledge}: Science fiction often explores the potential consequences of scientific and other innovations, and has been called a `literature of ideas'. \\ \textbf{History}: And that's difficult to do when dealing with time travel . I actually haven't seen the latest Harry Potter movies. Guess it's time to check them out!} & \textcolor{red}{If you really want a look at the potential negative consequences of scientific innovation, what you should check out is the TV show Fringe. Incredibly well written.}\\ \\ \\

        \midrule
        \textbf{CMU-DoG} &
        \multirow{5}{7cm}{\textbf{Knowledge}: Movie: The Social Network. In October 2003, 19-year-old Harvard University student Mark Zuckerberg is dumped by his girlfriend Erica Albright. Returning to his dorm, Zuckerberg writes an insulting entry about Albright on his LiveJournal blog and then creates a campus website called Facemash by hacking into college databases to steal photos of female students, then allowing site visitors to rate their attractiveness. After traffic to the site crashes parts of Harvard's computer network, Zuckerberg is given six months of academic probation. However, Facemash's popularity attracts the attention of Harvard upperclassmen and twins Cameron and Tyler Winklevoss and their business partner Divya Narendra. The trio invites Zuckerberg to work on Harvard Connection, a social network featuring the exclusive nature of Harvard students and aimed at dating.
        \\ \textbf{History}: The movie is The Social Network. I personally do not like Facebook as a company.}
        & The movie portrays the founding of social networking website Facebook and the resulting lawsuits. \textcolor{red}{It even has Justin Timberlake in it, I don't think I've ever seen him act.} \\ \\ \\  \\ \\ \\ \\ \\ \\ \\ \\ \\ \\ \\ \\ \\ 
        
         \midrule

        \textbf{TopicalChat} &
        \multirow{5}{7cm}{\textbf{Knowledge}: \underline{Wikipedia}: first paragraph in \url{https://en.wikipedia.org/wiki/Google}
        \underline{Reddit facts}: A single Google search requires more computing power than it took to send Neil Armstrong and eleven other astronauts to the moon. Google Maps calculates traffic by tracking how fast Android devices are moving on the road instead of hiring people to mow the lawns around their headquarters. Google uses hundreds of live goats. On 16th August 2013, Google went down for about five minutes, and took 40\% of web traffic with it. When there is a disputed border, Google maps tailors its maps to the claims of each country where the Internet browser is located.
        \\ \textbf{History}: Google provides online related services and products, which includes online ads, search engine and cloud computing.}
        & \textcolor{red}{Yeah, their services are good. I 'm just not a fan of intrusive they can be on our personal lives.} \\ \\ \\ \\ \\ \\ \\ \\ \\ \\ \\ \\ \\ \\ \\ \\
        
        \bottomrule
    \end{tabular}
    \caption{Hallucinated responses from different benchmarks: Wizard of Wikipedia \cite{dinan2018wizard}, \textsc{CMU-DoG} \cite{zhou-etal-2018-dataset} and \textsc{TopicalChat} \cite{Gopalakrishnan2019}. Text highlighted in red indicates hallucinated content.}
    \label{tab:hallu_gold_resp_3_datasets}
\end{table*}

\section{Expert Annotation}
\label{expert}
The two experts were students with linguistics background, fluent in English, and were trained for the task by exchanging rigorous discussions with the authors. 
 As part of this stage, they were required to write justifications for 50 samples articulating the reasoning for the provided ratings. The collected justifications were helpful in understanding the reasoning used to reach their ratings and in laying the groundwork for designing the second round of annotations.

\section{Inter-annotator Agreement on Gold Responses}
\label{kappa}
Table \ref{tab:kappa} contains the Fleiss kappa scores for \textsc{CMU-DoG} and \textsc{TopicalChat}.

\section{AMT Human Annotation}
\label{amt_ann}
\paragraph{Task Design}
To streamline the process for raters 
we break down the task into hierarchical (yes/no) questions. We summarize this procedure below, and provide the exact questions in~\S\ref{huma_ann_amt}. First, we ask annotators to judge whether the response contain information that is not supported by the source. If yes, we ask them to indicate the type of the unsupported information (e.g., unsupported opinion, unsupported fact, etc). In a followup question, we ask them to indicate whether there are any supported information besides the hallucinated content. If the response was not hallucinated, we present them with two follow-up questions about whether the response is entailing the source or generic. Finally, if the response entails the source, we ask whether it is coherent with the history. 
\paragraph{AMT Data Quality}
To access the initial staging round in AMT,  workers have to pass a qualification test by answering correctly 14 questions about \textsc{BEGIN} and \textsc{VRM}. Moreover, they had to be situated in the United States and Canada. Before being granted access to the main annotation task, 
workers would have access only to a small pilot round (batch size $\sim$ 50 HITs). In this round, we  carefully inspect each of the workers annotations for adherence to the instructions, and provide  feedback via email to those who committed errors.

At the end of this round, we revoke access for workers who provide poor quality annotations. Next, we launch the main annotation stage which is larger (batch size $\sim$ 400 HITs). We perform daily manual inspection and we send detailed feedback to workers who commit persistent error patterns. We reject poor quality work in this stage and repeated rejections lead to blocking the workers from the task indefinitely. In total, we ended up with 4 workers annotating the 4k responses. The workers were informed that their annotations would be used for research purposes and  their workers ID would be anonymous when we release the data.

\section{AMT Human Instructions}
\label{huma_ann_amt}
AMT Human annotation interfaces are depicted in Figure \ref{fig:amt1} and Figure \ref{fig:amt2}. We pay workers an hourly wage around 18-20 USD which is above the minimum wage rate.  Workers were asked the following questions:
 \begin{enumerate}
     \item Does the Wizard's response contain other information that is NOT supported by the evidence? (E.g., facts, opinions, feelings)?
     \begin{enumerate}
         \item If the response is hallucinated, what is the type of the unsupported information? (expressing a personal experience, expressing an opinion, expressing feelings, expressing unsupported facts, giving advice, acknowledging with information from the human)
         \item Besides unsupported information, does the Wizard's response contain thoughts/opinions/feelings/facts that are supported by the Evidence?

     \end{enumerate}

    \item  If the response is not hallucinated, is it faithful to the source or generic? (Faithful, Generic)
    
       \item If the response if faithful, is it cooperative with the Human's response?

 \end{enumerate}

\section{Limitation}
The main goal of this work is to present a data quality audit by gaining an in-depth understanding of the various types of hallucination in both gold and machine-generated responses. We do not investigate the root causes of hallucination in the models. Also, we limit our analysis to only English Benchmarks. Future studies can extend our work to explore the main causes of hallucination in the models and study the problem of hallucination in multilingual datasets.

\section{Hallucination in CMU-DoG and TopicalChat}
\label{hall_cmu_topi}
Figure \ref{fig:breakdown_gold_cmu_topical} shows the hallucination breakdown in \textsc{CMU-DoG} and \textsc{TopicalChat} benchamrks.

\section{Hallucinated Human-Human Responses}
\label{gold_hall_resp}
Table~\ref{tab:hallu_gold_resp_3_datasets} contains hallucinated gold responses from \textsc{WoW}, \textsc{CMU-DoG} and \textsc{TopicalChat}.

\section{Breakdown of BEGIN and VRM in Machine-generated Responses}
\label{app:vrm_ampl}
\begin{figure*}
    \vspace{-2em}
     \subfigure[\label{fig:wizard_gpt2}\footnotesize GPT2 responses]{\includegraphics[width=0.33\linewidth]{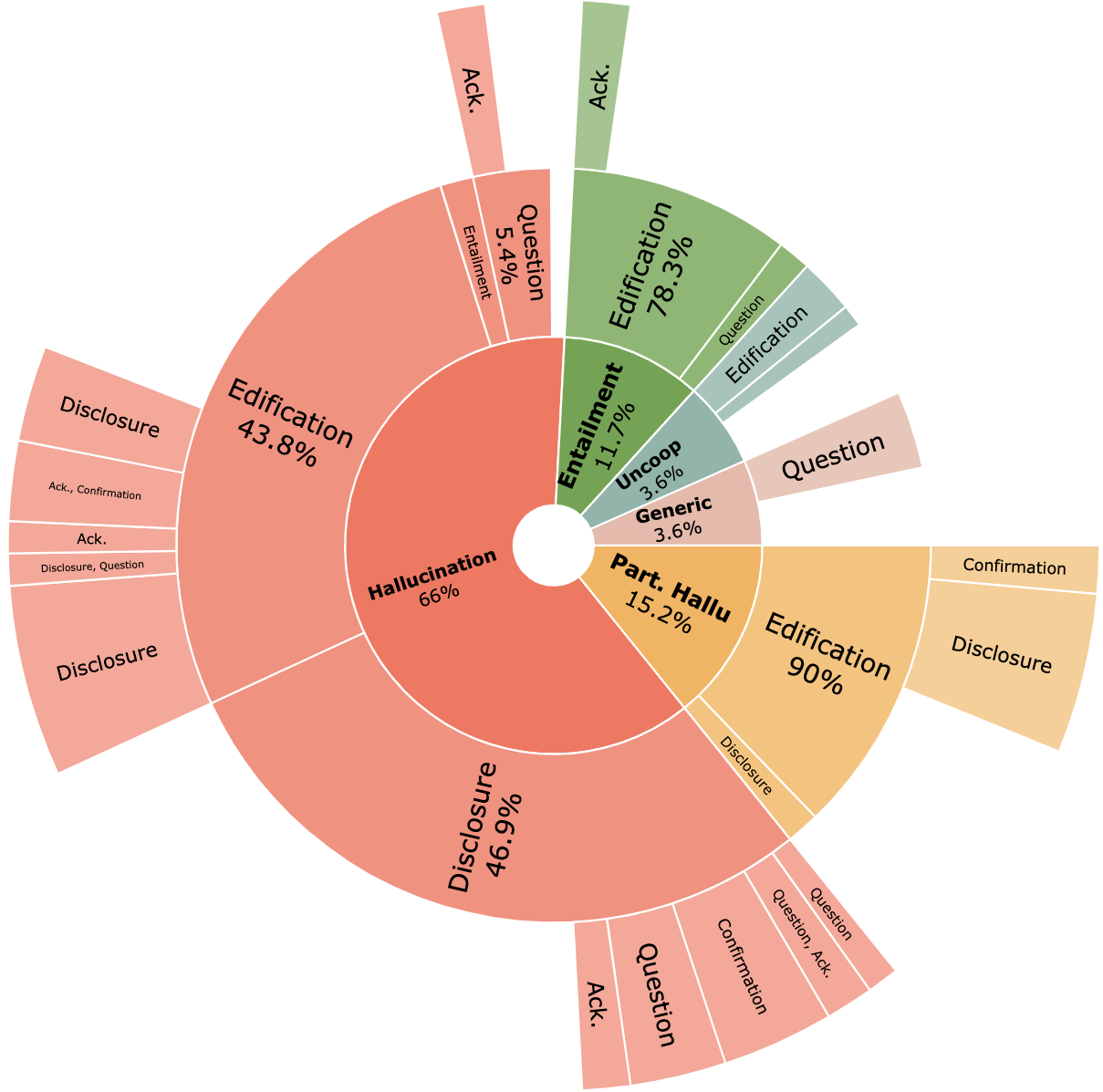}}
    \subfigure[\label{fig:wizard_doha}\footnotesize DoHA responses]{\includegraphics[width=0.33\linewidth]{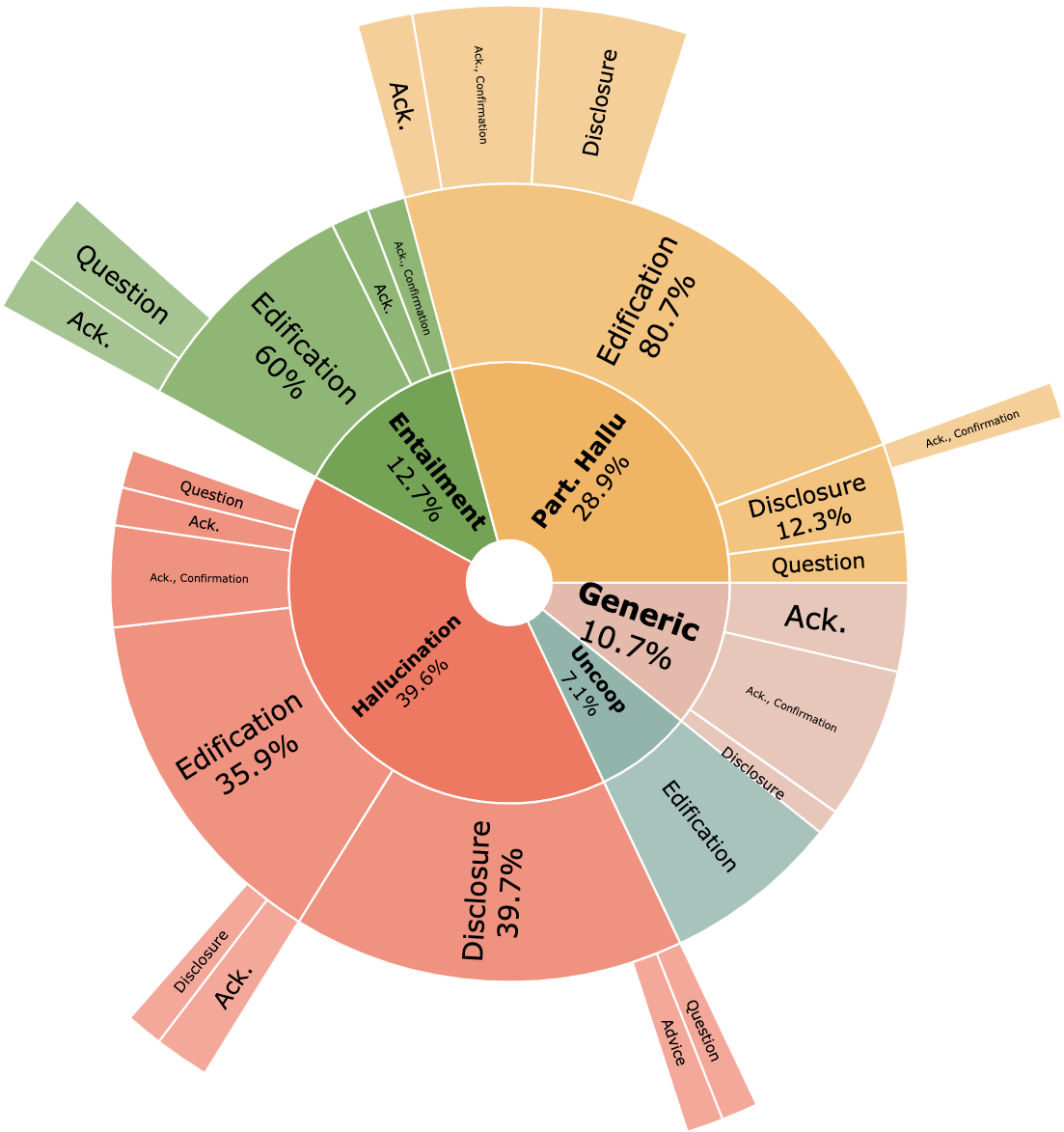}}
    \subfigure[\label{fig:wizard_ctrl}\footnotesize CTRL responses]{\includegraphics[width=0.34\linewidth]{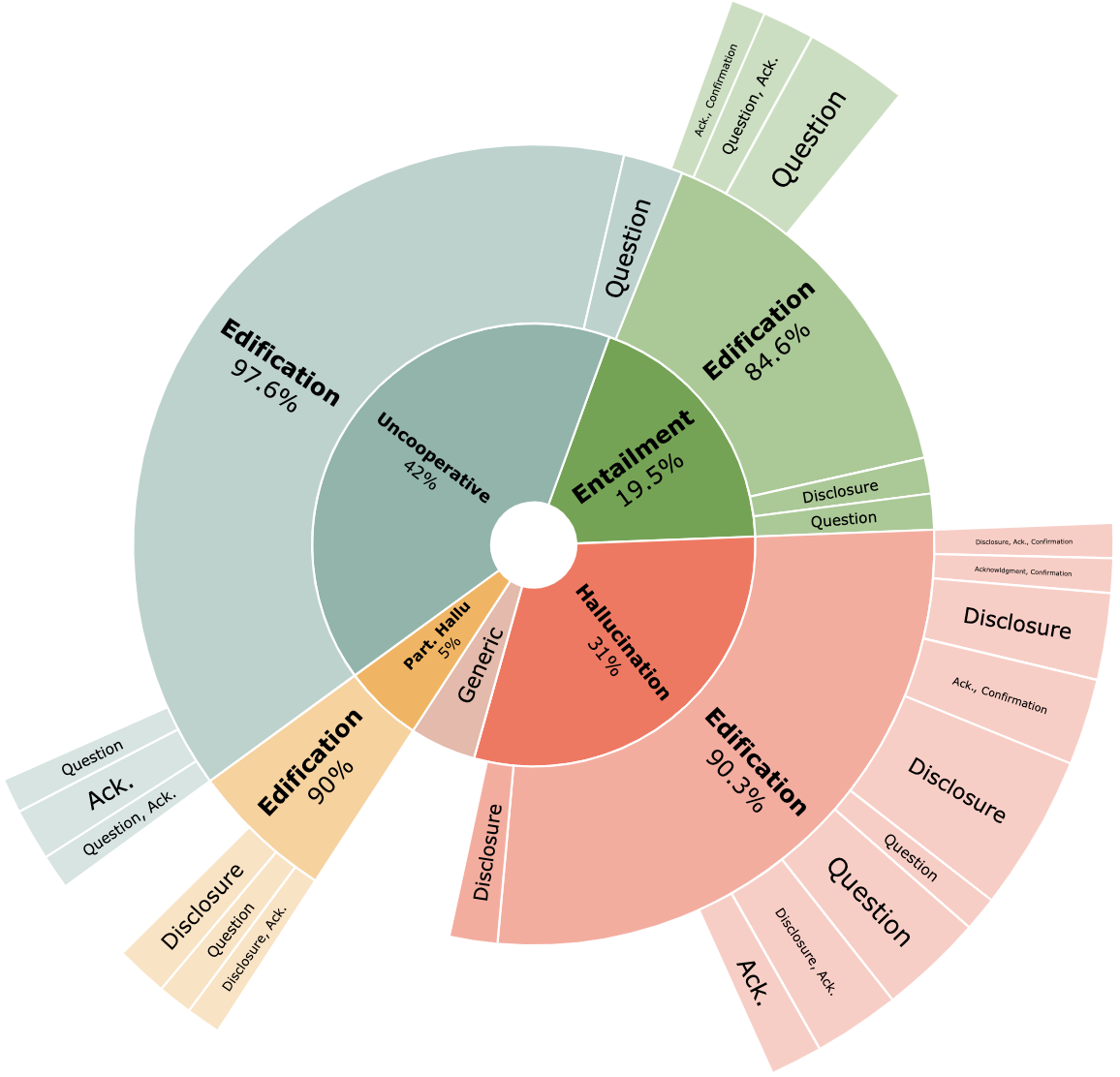}}
    \caption{Breakdown of BEGIN classes and VRM speech acts on WoW machine-generated responses.}
    \label{amplification-wow-new}
\end{figure*}

\begin{figure*}
    \vspace{-2em}
     \subfigure[\label{fig:cmu_gpt2}\footnotesize GPT2 responses]{\includegraphics[width=0.35
     \linewidth]{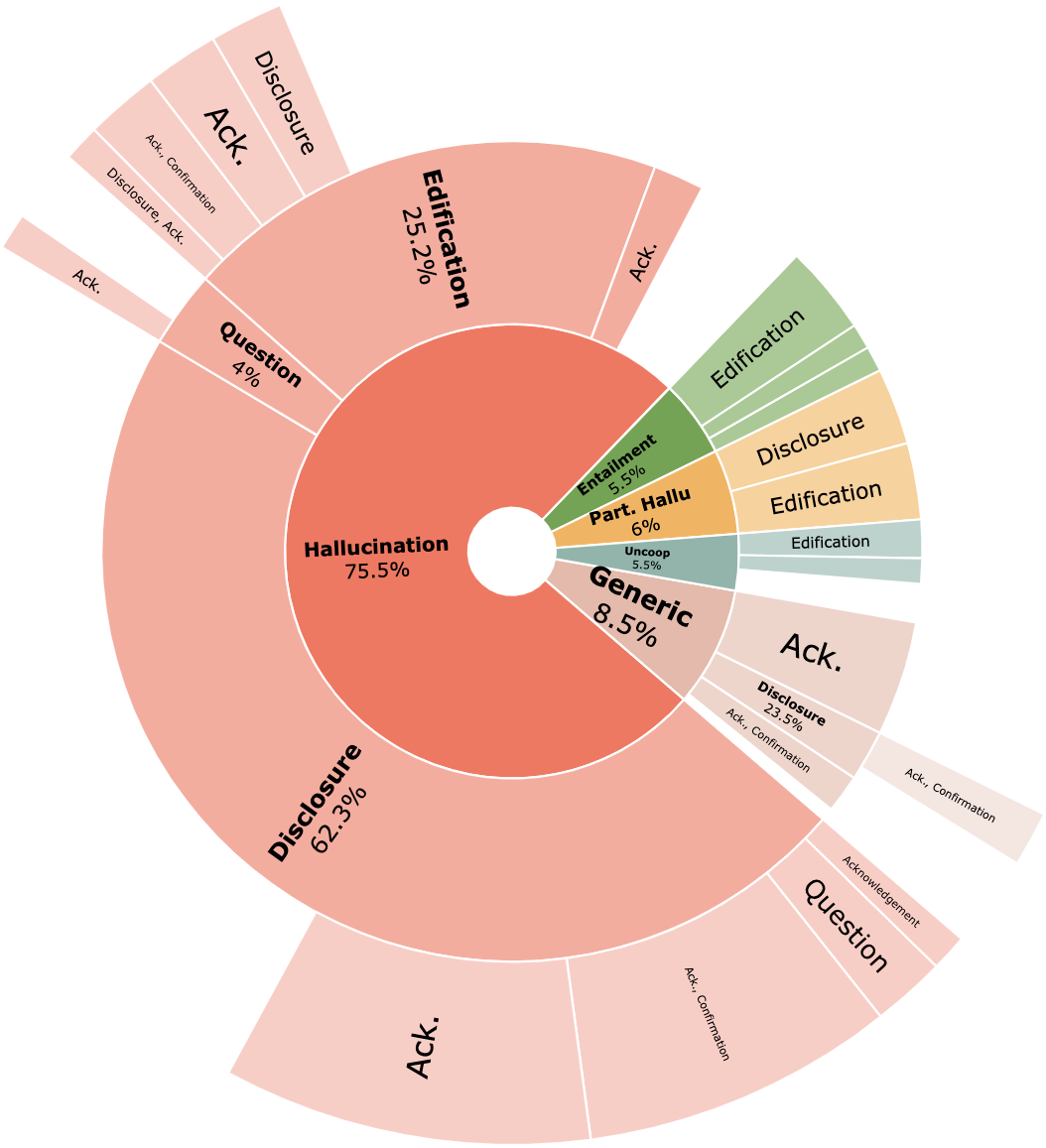}}
    \subfigure[\label{fig:cmu_doha}\footnotesize DoHA responses]{\includegraphics[width=0.31\linewidth]{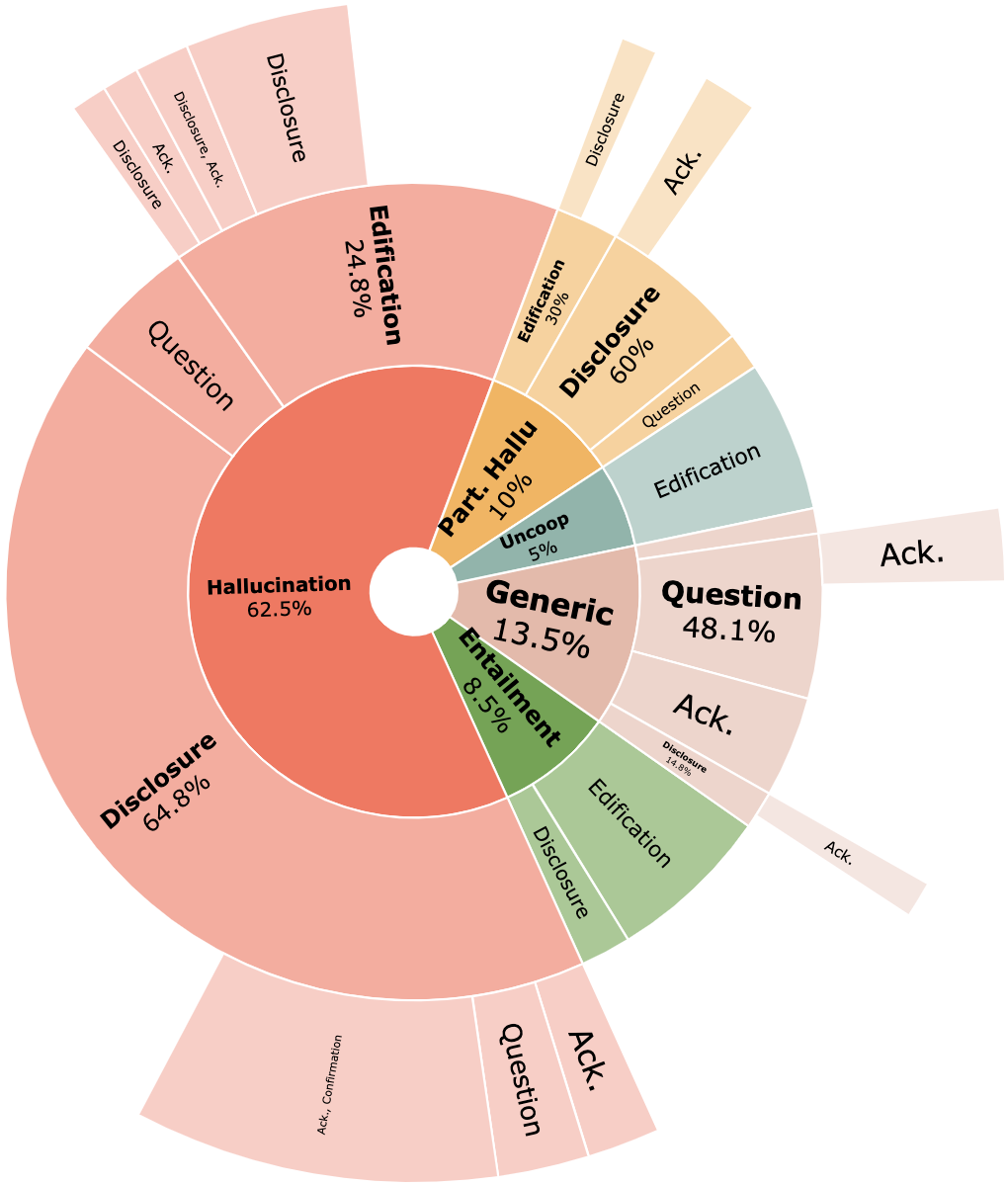}}
    \subfigure[\label{fig:cmu_ctrl}\footnotesize CTRL  responses]{\includegraphics[width=0.36\linewidth]{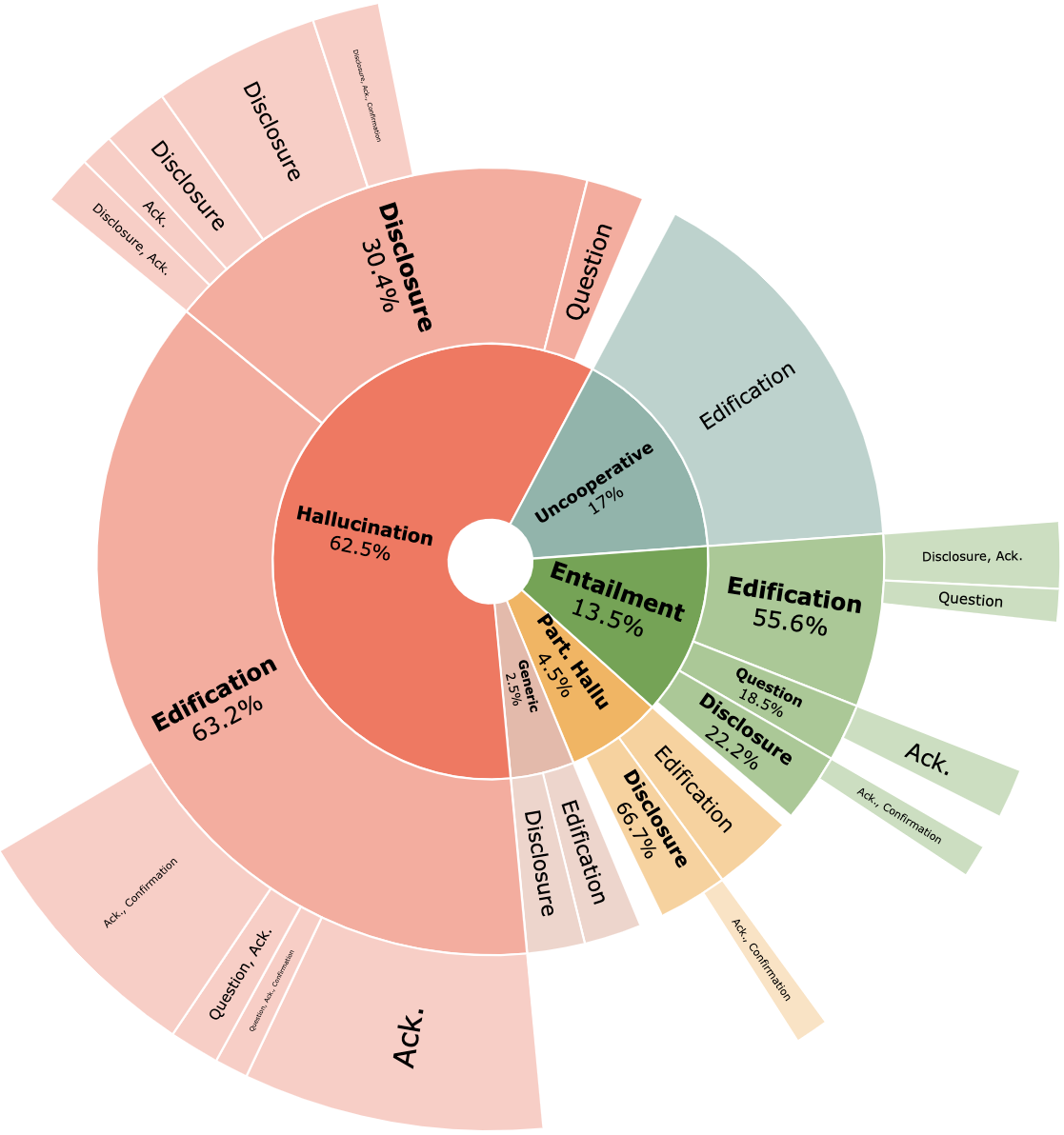}}
    \caption{Breakdown of \textsc{BEGIN} classes and \textsc{VRM} speech acts on \textsc{CMU-DoG} machine-generated responses.}
    \label{amplification-cmu-new}
\end{figure*}

\begin{figure*}
    \vspace{-2em}
     \subfigure[\label{fig:topical_gpt2}\footnotesize GPT2 responses]{\includegraphics[width=0.34
     \linewidth]{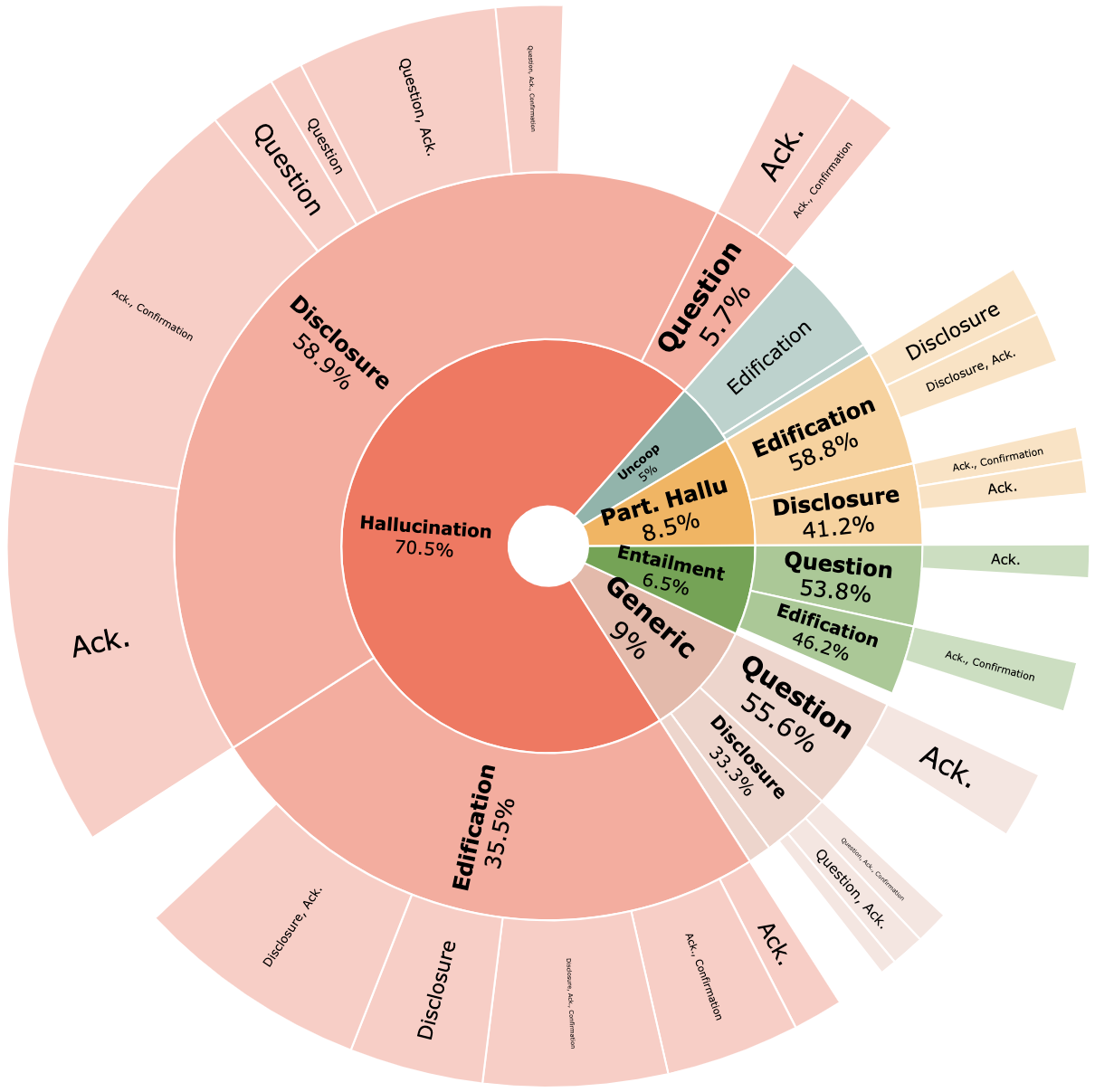}}
    \subfigure[\label{fig:topical_doha}\footnotesize DoHA responses]{\includegraphics[width=0.33\linewidth]{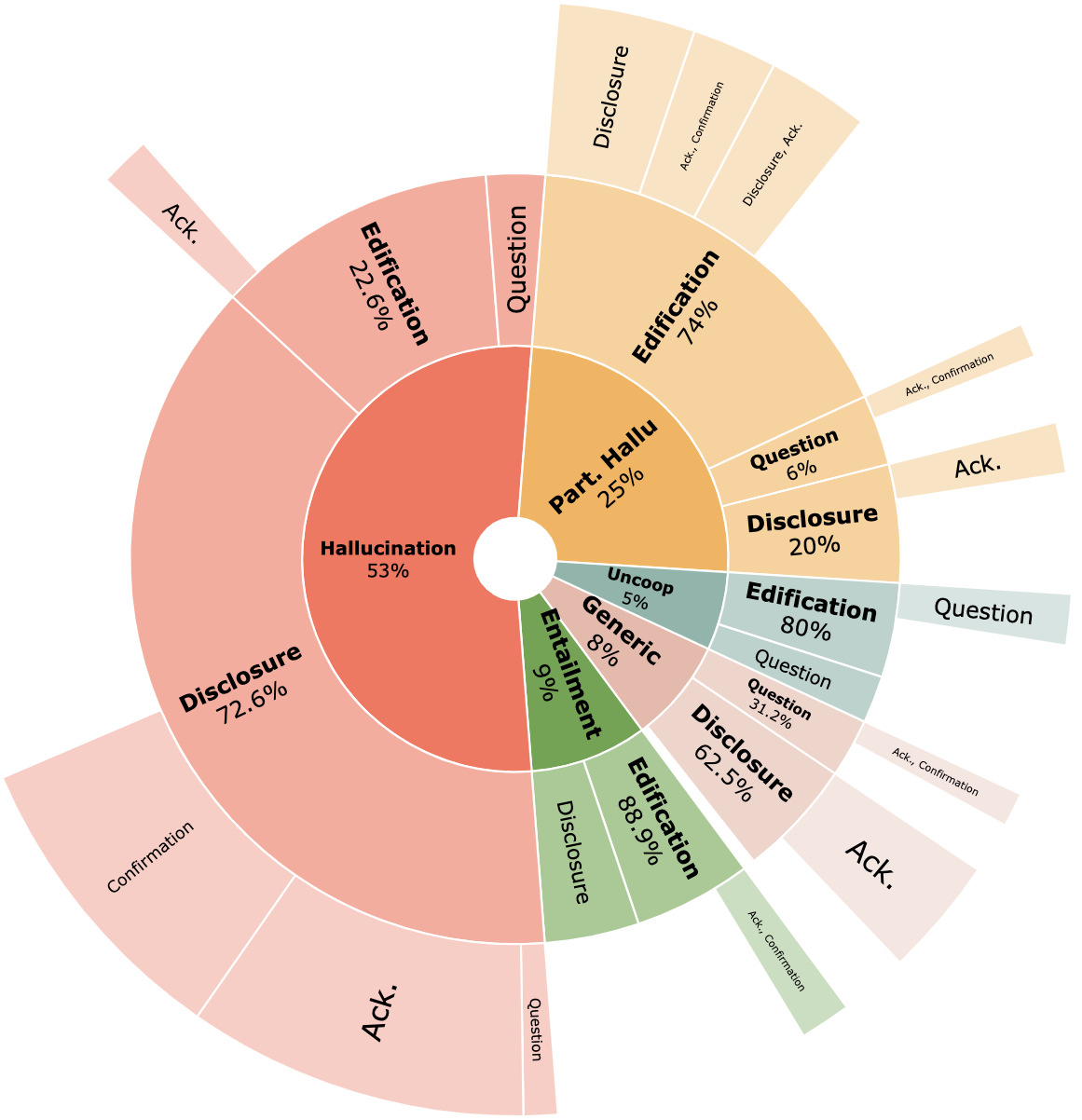}}
    \subfigure[\label{fig:topical_ctrl}\footnotesize CTRL responses]{\includegraphics[width=0.31\linewidth]{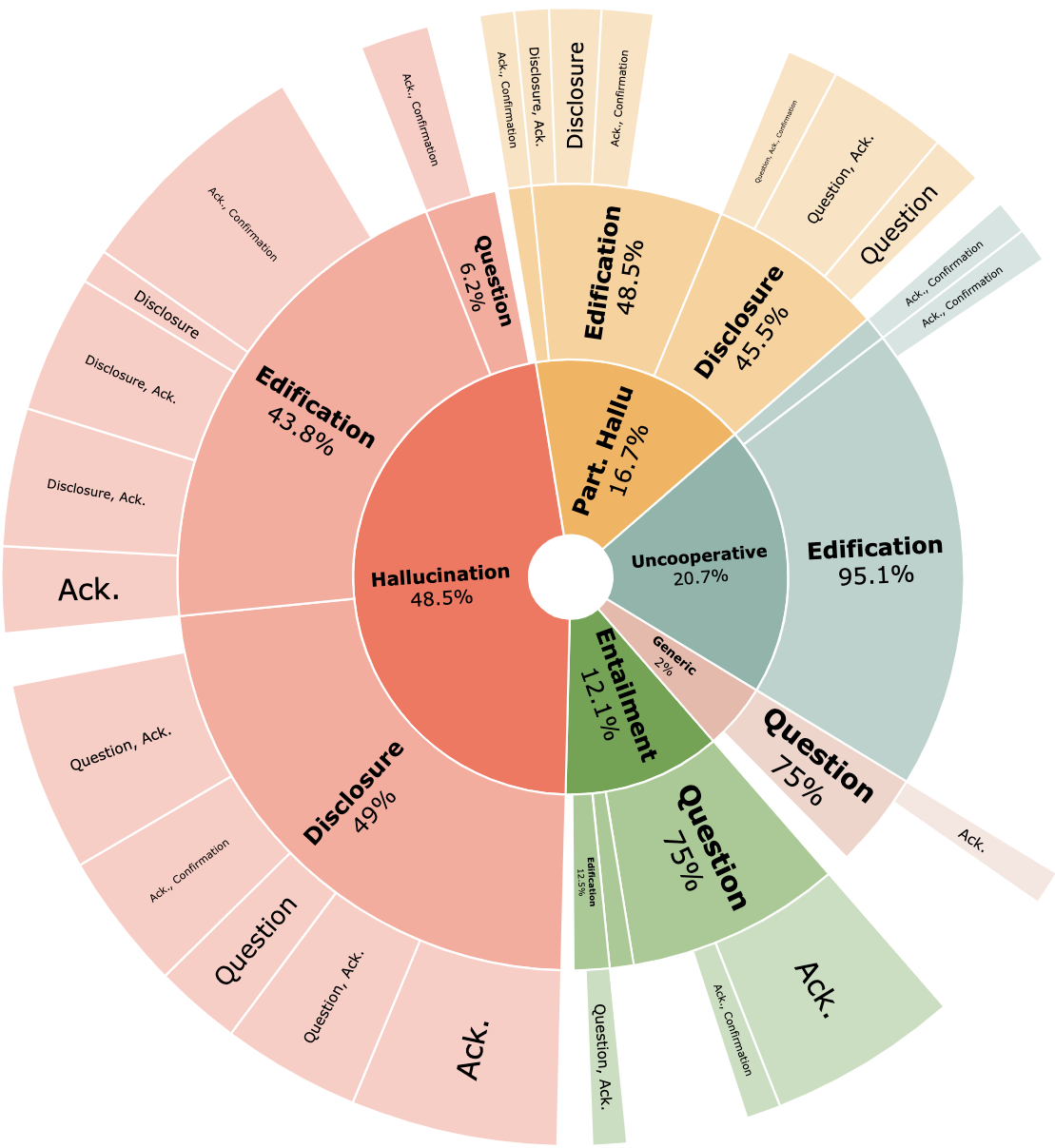}}
    \caption{Breakdown of BEGIN classes and VRM speech acts on Topical machine-generated responses.}
    \label{amplification-topical-new}
\end{figure*}

Figure \ref{amplification-wow-new}, \ref{amplification-cmu-new} and \ref{amplification-topical-new} display the distribution of \textsc{BEGIN} and \textsc{VRM} in \textsc{GPT2}, \textsc{DoHA} and \textsc{CTRL} trained on the three benchmark. 

\section{Machine-generated Responses}
\label{app:resp_output}
Table~\ref{tab:generated_Resp} contains a sample of generated responses from \textsc{GPT2}, \textsc{DoHA} and \textsc{CTRL} on the \textsc{WoW} and \textsc{CMU-DoG}.

\label{app:breakdown}

\end{document}